\documentclass[journal]{IEEEtran}
\usepackage{graphicx,amsfonts,stfloats,subfigure,color,multirow,booktabs,epsfig,pdfpages}
\usepackage{cite}
\usepackage[cmex10]{amsmath}
\usepackage{algorithmicx,algorithm,algpseudocode}
\usepackage{amssymb,bm}

\usepackage{hyperref}

\makeatletter
\adddialect\l@ENGLISH\l@english

\newcommand{\Rmnum}[1]{\expandafter\@slowromancap\romannumeral #1@}
\makeatother

\ifCLASSINFOpdf
\else
\fi
\begin{document}
%
\title{Orientation Driven Bag of Appearances for Person Re-identification}
%
%
%
\author{
        Liqian Ma,
        Hong Liu$^{\dag}$,~\IEEEmembership{Member,~IEEE,}
        Liang Hu,
        Can Wang,
        Qianru Sun
        \thanks{This work is supported by National Natural Science Foundation of China (NSFC, No.61340046, 60875050, 60675025), National High Technology Research and Development Program of China (863 Program, No.2006AA04Z247), Science and Technology Innovation Commission of Shenzhen Municipality(No.201005280682A, No.JCYJ20120614152234873, CXC201104210010A).}
        \thanks{L. Ma, H. Liu$^{\dag}$, L. Hu, C. Wang are with Key Laboratory of Machine Perception, Shenzhen Graduate School, Peking University, 518055 China (e-mail: maliqian@sz.pku.edu.cn; hongliu@pku.edu.cn; lianghu@pku.edu.cn; canwang@pku.edu.cn). Q. Sun is with
        Max-Planck-Institut f\"ur Informatik (e-mail: qianrusun@pku.edu.cn)}
        }

\maketitle

\begin{abstract}
Person re-identification (re-id) consists of associating individual across camera network, which is valuable for intelligent video surveillance and has drawn wide attention.
Although person re-identification research is making progress, it still faces some challenges such as varying poses, illumination and viewpoints.
For feature representation in re-identification, existing works usually use low-level descriptors which do not take full advantage of body structure information, resulting in low representation ability. 
To solve this problem, this paper proposes the mid-level body-structure based feature representation (BSFR) which introduces body structure pyramid for codebook learning and feature pooling in the vertical direction of human body.
Besides, varying viewpoints in the horizontal direction of human body usually causes the data missing problem, $i.e.$, the appearances obtained in different orientations of the identical person could vary significantly. To address this problem, the orientation driven bag of appearances (ODBoA) is proposed to utilize person orientation information extracted by orientation estimation technic.
To properly evaluate the proposed approach, we introduce a new re-identification dataset (Market-1203) based on the Market-1501 dataset and propose a new re-identification dataset (PKU-Reid). Both datasets contain multiple images captured in different body orientations for each person.
Experimental results on three public datasets and two proposed datasets demonstrate the superiority of the proposed approach, indicating the effectiveness of body structure and orientation information for improving re-identification performance.
\end{abstract}


%
\IEEEpeerreviewmaketitle

\section{Introduction}
%
%
%
%


\IEEEPARstart{P}{erson} re-identification deals with the recognition of individual who appears in non-overlapping camera views, which is fundamental and essential for intelligent video surveillance.
Generally, research difficulties lie in the ambiguity brought by the variations of poses, illumination and viewpoints. 

Recent years have witnessed lots of researches in this field. There are two major research aspects:
1) feature representation, including feature design \cite{BiCov,SDALF} and feature selection \cite{WhatFeaturesImportant,CUHK_SalienceMatching,ELF}; 2) model learning, including learning feature transform \cite{BTF} and learning distance metric \cite{ITML,KISSME} as introduced in the review article \cite{Re-IdChallenge}.
Considering feature representation, high-level features such as gender and age are difficult to reliably acquire due to the unconstrained viewpoints of individuals as well as the insufficiency of visual information in real-world surveillance scenarios.
Generally, most literatures describe body appearance with low-level descriptors \cite{SDALF,WhatFeaturesImportant,CUHK_SalienceMatching} which are usually sensitive to complex background and space misalignment.
In contrast, mid-level features could be more robust to serious space misalignment and can capture more discriminative vision information. Therefore, they could be robust to variations of poses and viewpoints \cite{Re-idLLC,CUHK_MidLevel,TCSVT15_reid_rgb,TCSVT15_reid_RD}.


Bag-of-Words (BoW) is a classical mid-level feature representation framework which has demonstrated excellent performances in computer vision tasks such as image classification \cite{SPM,LLC} and action classification \cite{LMY_ActionClassification,SQR_ActionClassification1,NC15_SQR}.
However, the traditional spatial pyramid widely used in BoW \cite{SPM,LLC} does not consider body structure information
which is important prior information for person re-identification.
Actually, individuals in images have roughly consistent structures in vertical direction, $e.g.$ head in the top and legs in the bottom. This observation allows us to describe the structure information of individuals using a common approach. Besides, different body parts have different color and texture characteristics, which suggests corresponding representations for different body parts.
Such prior knowledge provides richer information and leads to better re-identification performance \cite{SDALF,CI-DLBP}.
In this paper, we present a novel body-structure based feature representation (BSFR) approach for person re-identification.
A new body-structure pyramid is put forward to represent the body-structure information, meanwhile Locality-constrained Linear Coding (LLC) \cite{LLC}, one extension of BoW, is utilized to encode low-level descriptors into mid-level representations.
This BSFR method is a refined and expanded version of our conference paper \cite{BSFR}.

BSFR is used to describe a single-shot just like many previous works \cite{BiCov,SDALF,CUHK_SalienceMatching,ELF,WhatFeaturesImportant,Re-idLLC,CUHK_MidLevel}.
However, if people are tracked under each single camera, multi-shot person re-identification is more practical and may improve the results significantly since more information can be obtained. Typically, multi-shot methods can be divided into two groups: appearance based \cite{AHPE,Sarc3d,SDALF,Custom} and space-time based \cite{DTW4Re-id,VideoRank}.
With regard to the appearance based method, the appearance information from multiple frames can be fused in either feature level \cite{MRCG,AHPE,Sarc3d} or decision level \cite{SDALF,Custom}.
This paper follows the appearance based method and focuses on the feature level information fusion.
Comparing with decision level fusion, feature level fusion is capable of deriving and gaining the most effective and least dimensional feature vector sets that benefit the final decision \cite{FeaFusion}. In another respect, mid-level feature fusion is more robust to misalignment and background noise than low-level feature fusion. Besides, mid-level features are at a higher logical and semantical level than the low-level features, and mid-level features can avoid concatenating vectors of very different sizes \cite{TCSVT15_reid_rgbd}. Therefore, we fuse the multi-shot information with mid-level feature fusion.

\begin{figure}[t]
  \centering
  \includegraphics[width=8.7cm]{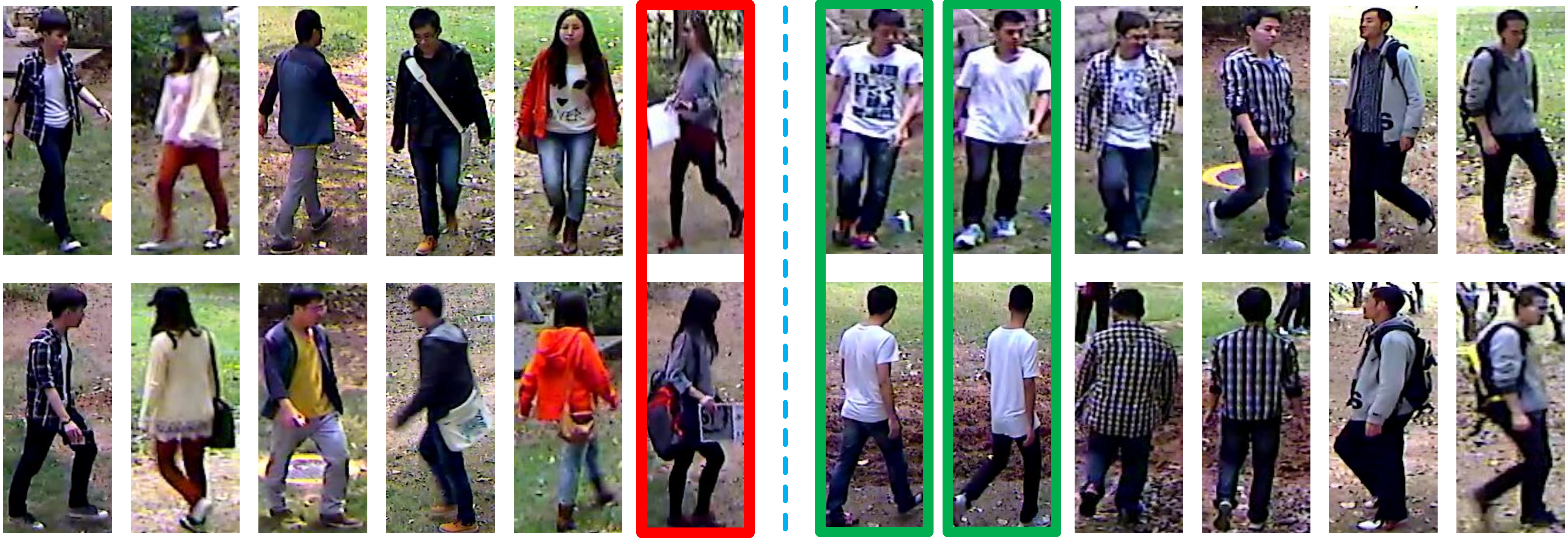} \\
  \vspace{-2mm}
  \caption{Left: the body appearances obtained in different orientations of the identical person may be dissimilar significantly. Right: appearances of different persons may be similar in some orientations (top line) but dissimilar in other orientations (bottom line). Each column corresponds to one person. \textbf{Best viewed in color.}}\label{fig_imgPairs}
\vspace{-3mm}
\end{figure}

Traditional methods are based on the hypothesis that the appearances of identical person are similar and the appearances of different persons are dissimilar, which may be invalid in some cases due to the varying viewpoints illustrated in Fig. \ref{fig_imgPairs}. For example, the female in red box shows very different appearances in different orientations, and appearances of the two males in green boxes are dissimilar in some orientation (top line) but similar in the other (bottom line).
Essentially, the variation of viewpoints is a data missing problem, because the appearances obtained in different orientations of the identical person could vary significantly, especially for the person with asymmetric clothes, bags and some other accessories.
Considering the varying viewpoints challenge, the body structure in horizontal direction, $i.e.$, the orientation information, is very helpful for multi-shot person re-identification.
However, most traditional methods directly fuse the multi-shot appearances with little or partial orientation information.
In this paper, we propose the ODBoA approach for multi-shot person re-identification. During the gallery set construction stage, ODBoA stores the multiple frames of the identical person in a bag according to the person orientation information.
During the matching stage, ODBoA selects the corresponding appearances from each person bag using person orientation information, and then constructs a single signature for each person with a max-pooling feature fusion strategy.
In addition, person appearances in one tracklet are more likely to have similar orientations.
Nevertheless, it is noted that the extension of camera network can produce relevant spatial-temporal constraint to obtain appearances of one person in dissimilar orientations with many differently oriented tracklets.
This important issue ensures that the approach proposed will yield an improvement \cite{Trans_Ori_Re-id}.

%

Taken together, both BSFR and ODBoA are constructed to make full use of body structure information in vertical direction and horizontal direction, respectively. In this paper, BSFR and ODBoA are combined to complement each other's advantages, $i.e.$,  multi-shot appearance model is constructed using ODBoA based on the single-shot feature representation extracted with BSFR.
Overall, this paper makes four major contributions,
\begin{itemize}
\item A body-structure based feature representation is introduced to describe the person in one image. It encodes low-level descriptors into a mid-level representation based on body structure information.
\item The ODBoA approach is proposed to describe a person with multiple images. It makes full use of person orientation information and works in two stages: 1) gallery set construction; 2) matching.
\item Two new person re-identification datasets, named PKU-Reid and Market-1203, are introduced.
    PKU-Reid dataset is composed of 114 identities collected by two cameras, and each person has eight images obtained in eight orientations in each scenario. To the best of our knowledge, PKU-Reid is unique that captures person images from eight equally divided orientations. Market-1203 dataset is constructed based on the Market-1501 dataset \cite{Market1501} and orientation is annotated for each image.
\item We fully evaluate the improvement in re-identification accuracy that can be attained by person orientation information. To be specific, we verify the great helpfulness of orientation information for metric model training and person matching stages. 
\end{itemize}

The rest of the paper is organized as follows. First, a brief review of the related works is provided in Section \Rmnum{2}. Then, the body-structure based feature representation is given in Section \Rmnum{3}, and the ODBoA approach is introduced in Section \Rmnum{4}. The datasets and evaluation protocol is presented in Section \Rmnum{5}. Experiments are presented in Section \Rmnum{6} and the conclusions are given in Section \Rmnum{7}.

\section{Related Works}
Feature representation is a core component in person re-identification. Typically, low-level descriptors such as color histogram and texture filters are used to describe person appearance \cite{BiCov,SDALF,WhatFeaturesImportant,KISSME,AHPE}. Su $et \ al.$ \cite{BiCov} design a novel BiCov feature to handle both background and illumination variations, which is based on the combination of biologically inspired features and covariance descriptors.
Gong $et \ al.$ \cite{WhatFeaturesImportant} combine 8 colour channels (RGB, HSV and
YCbCr) and 21 texture filters (8 Gabor filters and 13 Schmid filters) together and introduce a novel adaptive feature weighting method based on attribute-sensitive feature importance.
Koestinger $et \ al.$ \cite{KISSME} use low-level descriptors including HSV, LAB and LBP to describe person appearance, and match person with Mahalanobis distance metric learned by a simple but effective metric learning method.
Bazzani $et \ al.$ \cite{AHPE} design a novel HPE feature that incorporates complementary global and local statistical descriptions of the human appearance, focusing on the overall chromatic content via histogram representation, and the presence of recurrent local patches via epitomic analysis.

Considering that low-level descriptors may be more sensitive to space misalignment, some works focus on the mid-level feature representation.
Zhao $et \ al.$ \cite{CUHK_MidLevel} learn mid-level filters to represent features which could reach the balance between discriminative power and generalization ability, and achieve good performances. However, this method does not make full use of body structure information, and it has a high cost of computation.
Yang $et \ al.$ \cite{Re-idLLC} introduce LLC to encoded low-level descriptors into mid-level features, which has a low cost of computation and better discrimination but does not consider body structure information either.
To employ the recent advances of Fisher Vectors for person re-identification, Su $et \ al.$ \cite{LDFV} introduce a new Local Descriptors encoded by Fisher Vector (LDFV) descriptor to encode local features into a global vector.
Zheng $et \ al.$ \cite{Market1501} consider the re-identification task as a ¡°query-search¡± problem and apply image search technics such as BoW model and TF-IDF scheme to person re-identification.
These methods mentioned above usually use equally divided horizontal strips as the geometric constraints of human body,
which considers little human body structure information.
To utilize more body structure information, Bazzani $et \ al.$ \cite{SDALF,AHPE} propose an adaptive body segmentation approach for re-id based on foreground segmentation technics. However, foreground segmentation is very sensitive to complex background, thus the segmentation may vary obviously for the identical person in different scenarios or orientations. 
Cheng $et \ al.$ \cite{Custom} apply Pictorial Structures (PS) to segment body parts for single-shot re-identification and develop Custom Pictorial Structures (CPS) for multi-shot re-identification. However, Custom PS (CPS) is a two-step iterative process, and it is time-consuming. 
Taking both body structure representation and computing speed into account, we propose an effective and efficient mid-level feature representation approach called BSFR for a single image to introduce body structure information for codebook learning and feature pooling.

\begin{figure}[t]
  \centering
  \includegraphics[width=8.7cm]{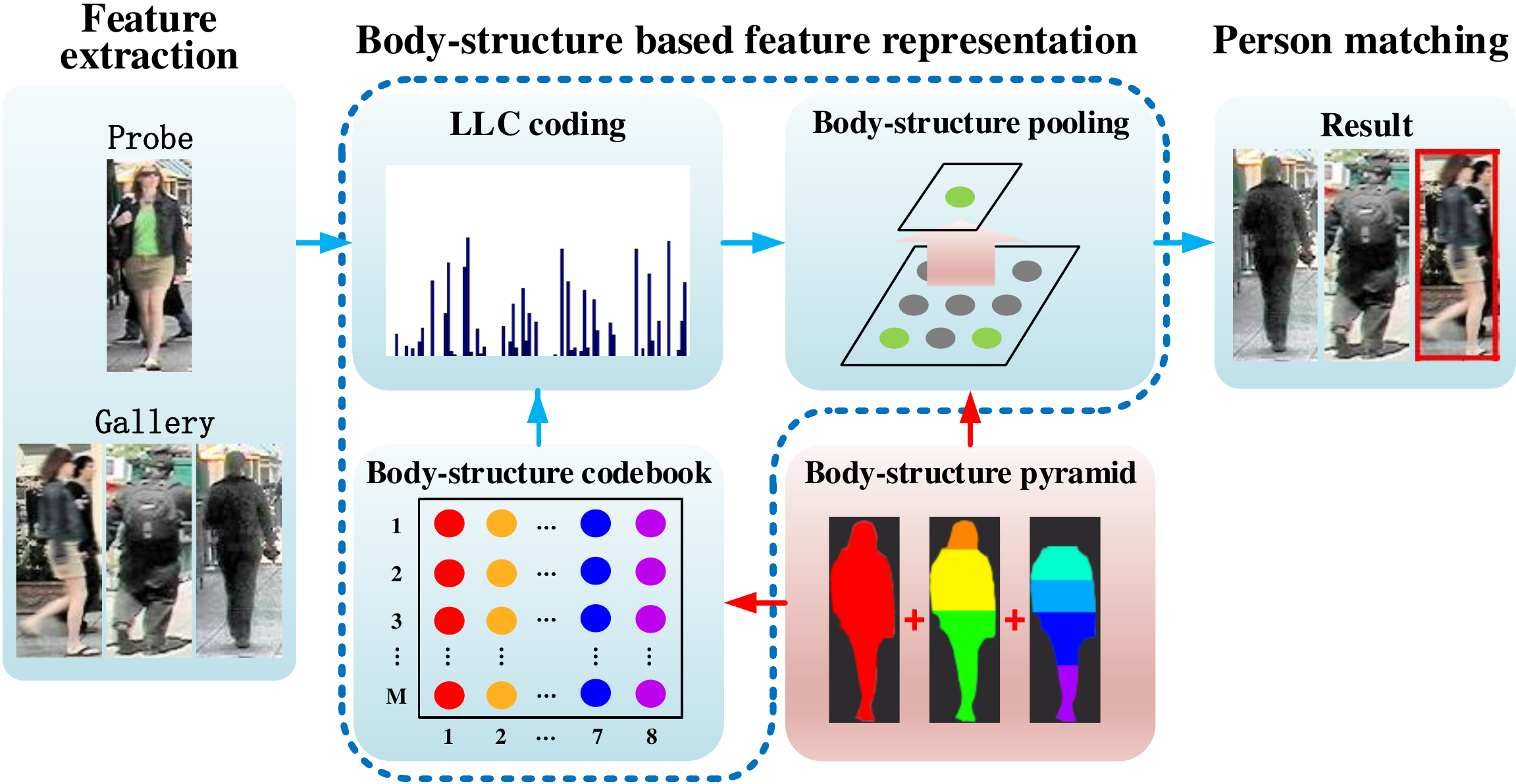} \\
  \vspace{-2mm}
  \caption{The pipeline of body-structure based feature representation. The eight distinct colors in the body-structure pyramid are related to eight sub-codebooks in body-structure codebook. \textbf{Best viewed in color.}}\label{fig_pipeline}
\vspace{-3mm}
\end{figure}

Typically, multi-shot person re-identification can utilize more information to improve matching accuracy. 
One way to fuse multiple frames appearance information is feature level fusion.
To obtain highly discriminative human signature, Bak $et \ al.$ \cite{MRCG} propose Mean Riemannian Covariance Grid (MRCG) to combine information from multiple images.
Bazzani $et \ al.$ \cite{AHPE} condense a set of frames of an individual into a highly informative signature via histogram representation and epitomic analysis based on foreground extraction.
Alavi $et \ al.$ \cite{RDC} represent each image with a modified manifold technic and employ nearest neighbour for final classification.
Baltieri $et \ al.$ \cite{Sarc3d} propose a new simplified 3D body model computed from 2D person images detected and tracked in each calibrated
camera. While the construction of 3D body model relies on camera calibration, precise foreground extraction and 3D orientation estimation technics which may be unstable with complex backgrounds.
%
The other way is decision level fusion. The straightforward idea is to calculate the distance of each image pair and use the average or minimal one as the final distance between two persons \cite{SDALF,Custom}.
Wu $et \ al.$ \cite{Collaborative_multi-shot} apply set-based matching to multi-shot person re-identification with collaborative sparse approximation, which does not consider person orientation information.
Garcia $et \ al.$ \cite{Trans_Ori_Re-id} rank the multi-shot re-identification result with orientation information based on single-shot pairwise distance. Further more, they propose a dual-classification method \cite{ICPR14_Ori_Re-id} to calculate pairwise feature dissimilarities with different classifier based on orientation distance. 
We point out that in these works the improvement in re-identification accuracy that can be attained by orientation information has not been clearly evaluated.

Oliver $et \ al.$ \cite{Re-idInWild} introduce the concept of bag of appearances (BoA) which is a container of color features that fully represents a person by collecting all his different appearances obtained from Kinect. They perform person matching in a probabilistic framework by accumulating the probability of pairwise matching for all of the elements in each bag with appearance and height information.
However, BoA contains much redundant data redundancy and ignores the orientation information, resulting in limited accuracy, large storage cost and computation cost.
Inspired by the concept of BoA \cite{Re-idInWild}, we introduce ODBoA to store and select the candidate elements in each bag for person matching.
Since mid-level feature fusion describes person appearance more comprehensively and is more robust to misalignment and background noise, a mid-level feature pooling strategy is employed to construct a single signature for each person based on BSFR.

\section{Person Feature Representation}
In this section,
we propose an approach to encode low-level descriptors into mid-level features using body structure information. As depicted in Fig. \ref{fig_pipeline}, feature representation based on body-structure is performed after feature extraction, followed by person matching stage. For BSFR, human body is first split into eight parts according to body structure information \cite{BodyParts} to construct the body-structure pyramid. Then, it is used as reference information to learn the body-structure codebook and pool the features encoded by LLC.

\vspace{-2mm}
\subsection{Body-structure pyramid}
The substantial body structure information of people is very helpful for re-identification. Our body-structure pyramid is designed based on the following three observations: (1) Vertical space misalignments caused by pose and viewpoint variations appear much less than horizontal space misalignments; (2) Human body is not a rigid object for its complex kinematics, so it can be better described using a part-based model; (3) Spatial layout information is considerably critical information and can be used to describe body appearance.

\begin{figure}[t]
  \centering
  \hspace{-2mm}
  \includegraphics[width=8.5cm]{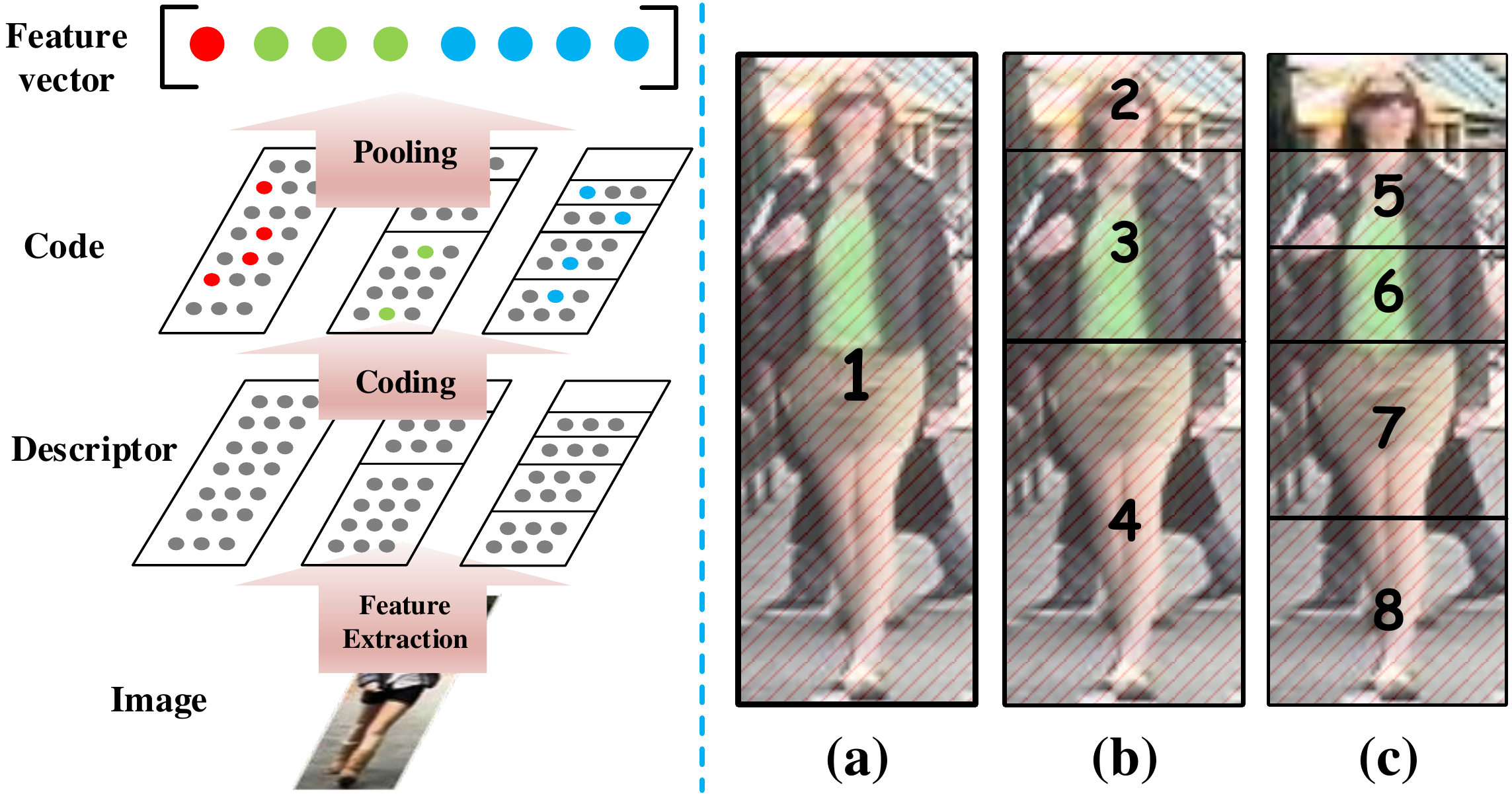}
  \vspace{-1.3mm}
  \caption{Left: flowchart of the body-structure pyramid for pooling features. Right: our proposed body-structure pyramid composed of eight parts from three levels. (a) One horizontal strip of whole body. (b) Three horizontal stripes of 16\%, 29\% and 55\% of the total pedestrian height locate head, torso and legs \cite{BodyParts}, respectively. (c) Four horizontal stripes built by subdividing torso part and leg part into two subparts with equal size. The head part in (c) is as same as the one in (b). \textbf{Best viewed in color.}}\label{fig_pyramid}
  \vspace{-2.5mm}
\end{figure}

Adaptive part models based on background substraction technics are used in some previous works \cite{DCD,SDALF,Custom} and have gained some performance enhancements in certain situations.
However, these adaptive part models require more computation cost and may generate incorrect segmentation in complex scenarios.
In this paper, motivated by \cite{BodyParts,SPM}, body structure information is utilized to construct the body-structure pyramid as shown in Fig. \ref{fig_pyramid}(a)(b)(c).
An effective fixed part model is proposed to describe body appearance. It solves the space misalignment by dividing the pedestrian image into increasingly fine vertical sub-regions with some prior knowledge of body structure.
As depicted in Fig. \ref{fig_pyramid}(b) three horizontal stripes of 16\%, 29\% and 55\% of the total pedestrian height respectively locate head, torso and legs \cite{BodyParts}. Further, torso part and leg part are both subdivided into two horizontal stripes with equal size as shown in Fig. \ref{fig_pyramid}(c), so as to describe human body in a finer level. The total eight parts in Fig. \ref{fig_pyramid}(a)(b)(c) compose the body-structure pyramid.

\subsection{Body-Structure based Feature Representation}

\subsubsection{Body-structure codebook learning}
As different body parts have different characteristics, body-structure pyramid is used to construct a body-structure codebook in order to increase the discrimination of codebook.
The patches, sampled from images, are divided into eight patch sets according to the body-structure pyramid,

\vspace{-2mm}
\begin{eqnarray}\label{eq1}
    P_a=\{p_{j,t} | p_{j,t} \in r_a, t=1,...,T \}
\end{eqnarray}
where $P_a$ denotes the $a$-th patch set and $p_{j,t}$ denotes the $j$-th patch of the $t$-th image, while $r_a$ is the $a$-th part of body-structure pyramid.
K-means is applied to learn each sub-codebook using the descriptors extracted from patches randomly selected from the relevant patch set $P_a$. The final body-structure codebook consists of eight sub-codebooks as shown in Fig. \ref{fig_pipeline}, and each sub-codebook has $M$ entries with $D$ dimensions,
\begin{equation}\label{eq2}
    \begin{aligned}
        &B=\{B_a | a=1,...,N\}\\
        B_a&=[b_{a,1},b_{a,2},...,b_{a,M}] \in \mathbb{R}^{D \times M}
    \end{aligned}
\end{equation}
where $B$ is the body-structure codebook and $B_a$ is the $a$-th sub-codebook. $N$ denotes the number of sub-codebook and $M$ denotes the number of entities in each sub-codebook.

\subsubsection{LLC coding}
LLC \cite{LLC} is a fast and effective feature coding method applied to image classification task successfully.
In this paper, LLC is adopted to encode mid-level features using body-structure codebook as shown in Fig. \ref{fig_pipeline}. LLC gives an analytical solution for the following criteria,
\begin{equation}\label{eq3}
    \begin{aligned}
            \min_{\substack{C}} \sum_{i=1}^N
                &{\lVert x_i - B_a c_i \rVert}^2 + \lambda {\lVert d_i \odot c_i \rVert}^2 \\
            &s.t. 1^\top c_i = 1, \forall i
    \end{aligned}
\end{equation}
where $\odot$ denotes the element-wise multiplication, and $d_i \in \mathbb{R}^M$ is a locality adaptor with different proportion for each basis
according to its similarity to the input descriptor $x_i$,
\begin{eqnarray}\label{eq4}
    \begin{aligned}
        d_i=exp\begin{pmatrix}\cfrac{dist(x_i,B_a)}{\sigma}\end{pmatrix} \\
    \end{aligned}
\end{eqnarray}
where $dist(x_i,B_a)=[dist(x_i,b_{a,1}),...,dist(x_i,b_{a,M})]$, and $dist(x_i,b_{a,j})$ is the Euclidean distance and $\sigma$ is used to adjust the weight decay speed \cite{LLC}.
Regularization term in Eq. \eqref{eq3} leads to locality, which can generate similar codes for similar descriptors and make the features more discriminative.
Further more, the work in \cite{LLC} gives an approximated LLC for fast coding, which reduces the computation complexity significantly.

\begin{figure}[t]
  \centering
  \includegraphics[width=8.5cm]{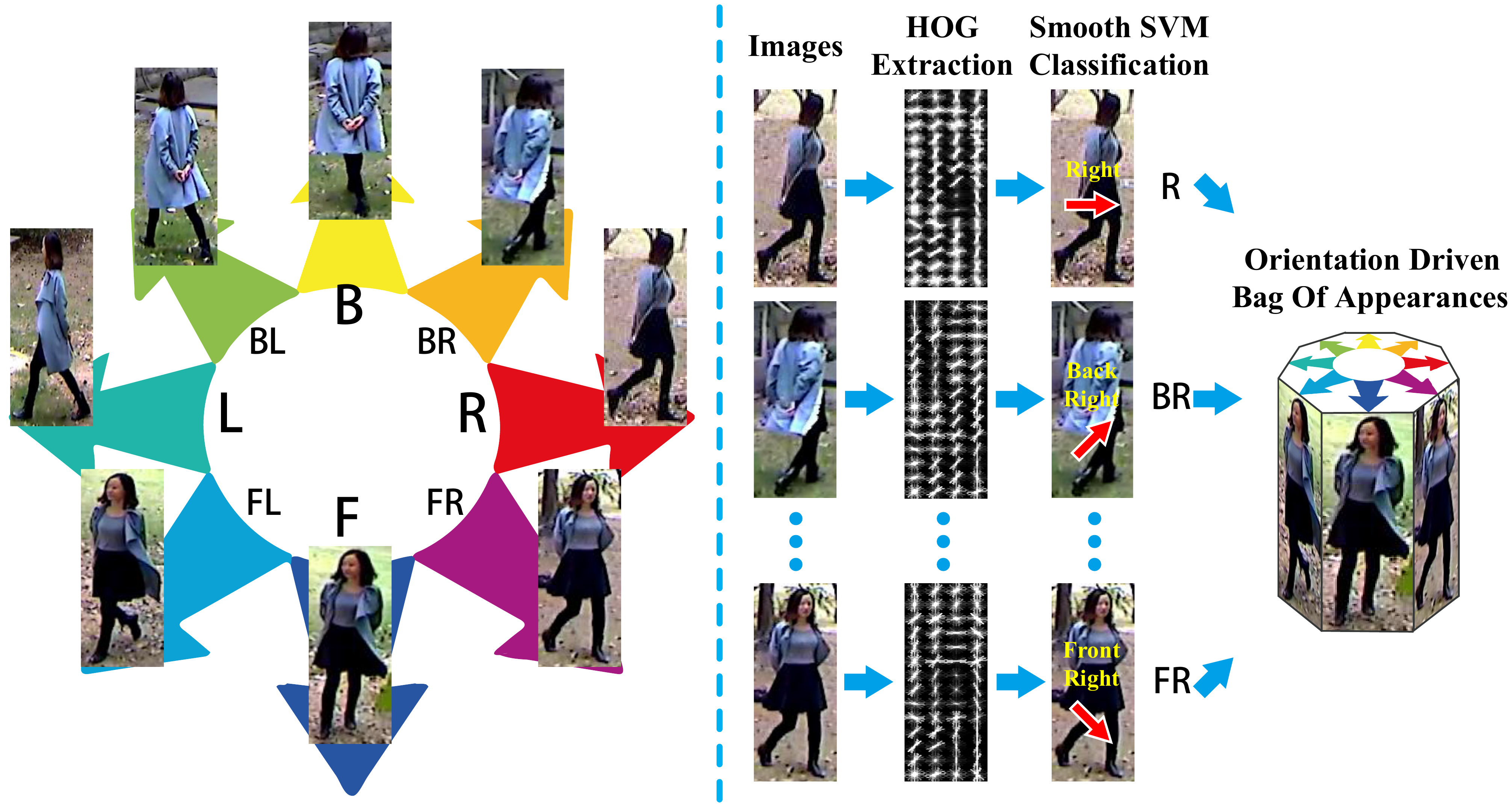}
  \vspace{-1.8mm}
  \caption{Left: illustration of eight orientations used in our framework. Right: the construction process of ODBoA. \textbf{Best viewed in color.}}\label{fig_ori}
  \vspace{-2.5mm}
\end{figure}

\subsubsection{Body-structure pooling}
Feature pooling is an effective way to select features and can achieve some invariance of space misalignment.
As shown in Fig. \ref{fig_pyramid}, a feature pooling strategy using body-structure pyramid as reference information is proposed to incorporate body structure information into the feature representation well.
Body-structure pooling combines the codes of the same body part into a single feature vector and makes the feature vector invariant to person space misalignment, especially the horizontal one caused by varying poses and viewpoints.
Since max pooling over sparse codes is robust to clutter \cite{Mid-levelFeatures} and
can capture the salient properties of local regions \cite{ScSPM}, we uses max pooling,
\begin{equation}\label{eq5}
    \begin{aligned}
    f_{a} = max(c_{a,1},c_{a,2},...,c_{a,K})
    \end{aligned}
\end{equation}
where ``$max$'' function runs in a row-wise manner, pooling codes $c_{a,i}$ in the $a$-th part of body-structure pyramid into one feature vector $f_{a}$, and $K$ denotes the number of
descriptors in this part.
Finally, feature representation is obtained by concatenating then $\ell^2$ normalizing the pooled features.

\begin{figure}[t]
  \centering
  \hspace{-3mm}
  \includegraphics[width=8cm]{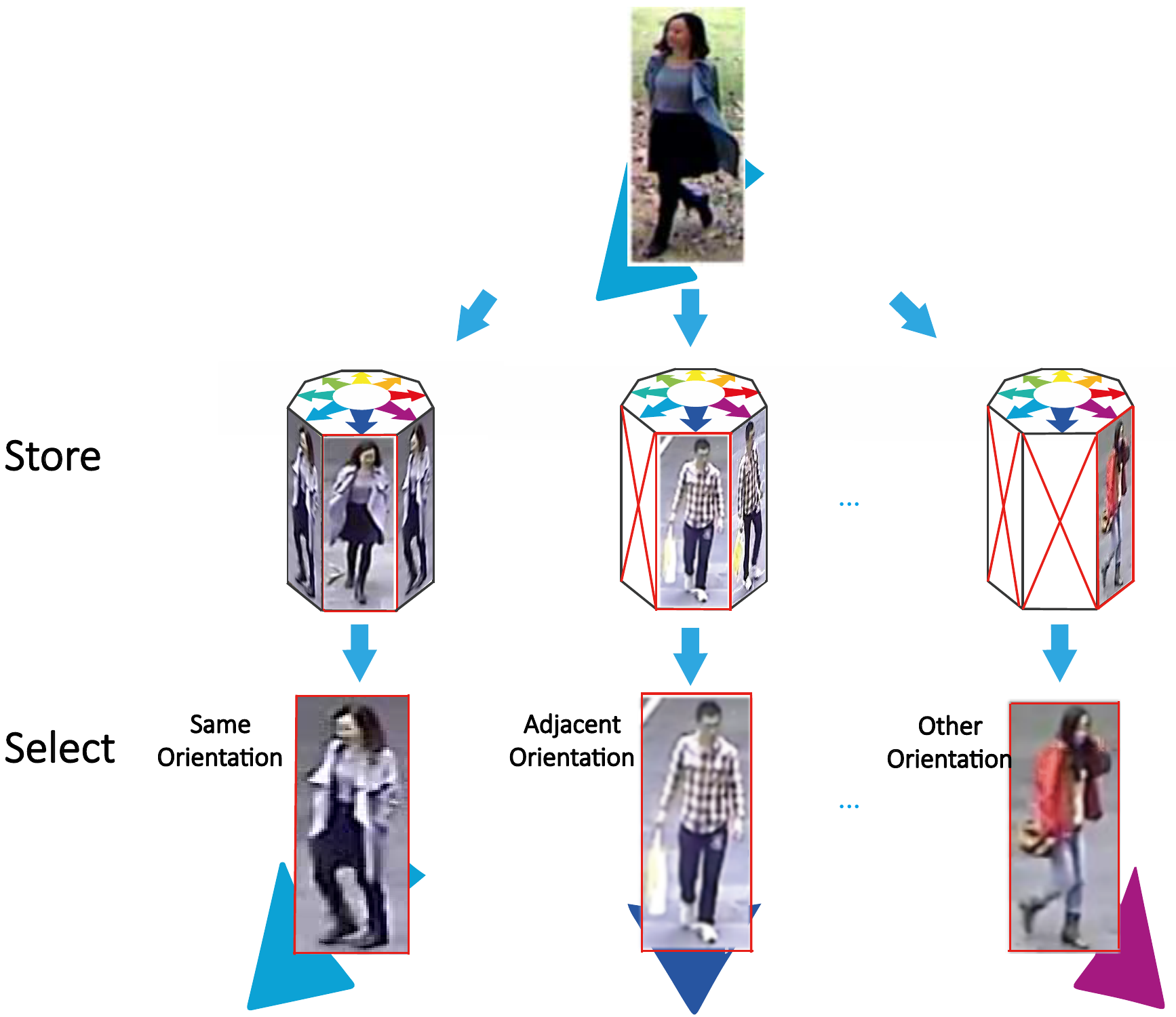}
  \caption{The orientation based storage and selection process of ODBoA. \textbf{Best viewed in color.}}\label{fig_ODBoA}
\end{figure}

\section{Proposed Framework for Person re-identification}
After introducing the body-structure based feature representation, this section provides a detailed description on how to perform multi-shot person re-identification.
First, person image feature representation is calculated using BSFR mentioned in Section \Rmnum{3}. Second, the concept of ODBoA is introduced to describe a person. Then, the mid-level features are matched through a Mahalanobis distance metric learned by Keep It Simple and Straightforward MEtric (KISSME) \cite{KISSME} which is an efficient metric learning method \cite{Trans_rskissme, Trans_mcekissme, SCNCD, CVPR15_LOMO}.



\subsection{Orientation driven bag of appearances}
Considering that appearances obtained in different orientations of the identical person could vary significantly, especially for person with asymmetric clothes, bags and some other accessories.
An ODBoA is a container of person appearances obtained in different orientations. It is used to store candidate frames then to select the suitable ones for matching.

\subsubsection{ODBoA construction}
Following \cite{SemiOri}, we consider eight quantized orientations: Right (R), Back-Right (BR), Back (B), Back-Left (BL), Left (L), Front-Left (FL), Front (F), Front-Right (FR), as illustrated in Fig. \ref{fig_ori}.  
To estimate person orientation, a baseline method is applied.
First, Histograms of Oriented Gradients (HOG) descriptor is employed to capture the local shape information. Second, linear SVM with the one-vs-all scheme is used to train the eight classifiers for each orientation. Finally, the probabilities obtained from the classifiers are smoothed in order to integrate the classification abilities of these classifiers, since the overlapping of the orientation classes leads to more than one high response from the set of discrete-orientation classifiers due to the continuity of angle \cite{ECCV12_Ori}.
The smooth strategy is as follows,
\begin{equation}\label{eq6}
    \psi_{i} = \sum_{k=-1}^{+1}w_{k}\cdot\psi_{\theta(k)}\\
\end{equation}
\begin{equation}\label{eq7}
    \theta(k) = ((i+k)-1)mod8+1
\end{equation}
where $\psi$ and $w$ are the probability and weight, and $i\in{\{1,2,...,8\}}$ denotes eight discrete orientations.
Finally, the appearances obtained in different orientations are integrated into a ODBoA model to describe the person appearance information as shown in Fig. \ref{fig_ori}.

For each person we construct one bag. If there exist multiple frames of one person in the same orientation, the feature vectors obtained from these frames will be pooled into one vector with max pooling to represent the appearances in this orientation, since these frames may contain different information. It reduces the data redundancy, but still retains the information of different frames. Hence, in one bag, each orientation corresponds to a single feature vector.


\subsubsection{ODBoA matching}
\begin{figure}[t]
  \centering
  \includegraphics[width=8.5cm]{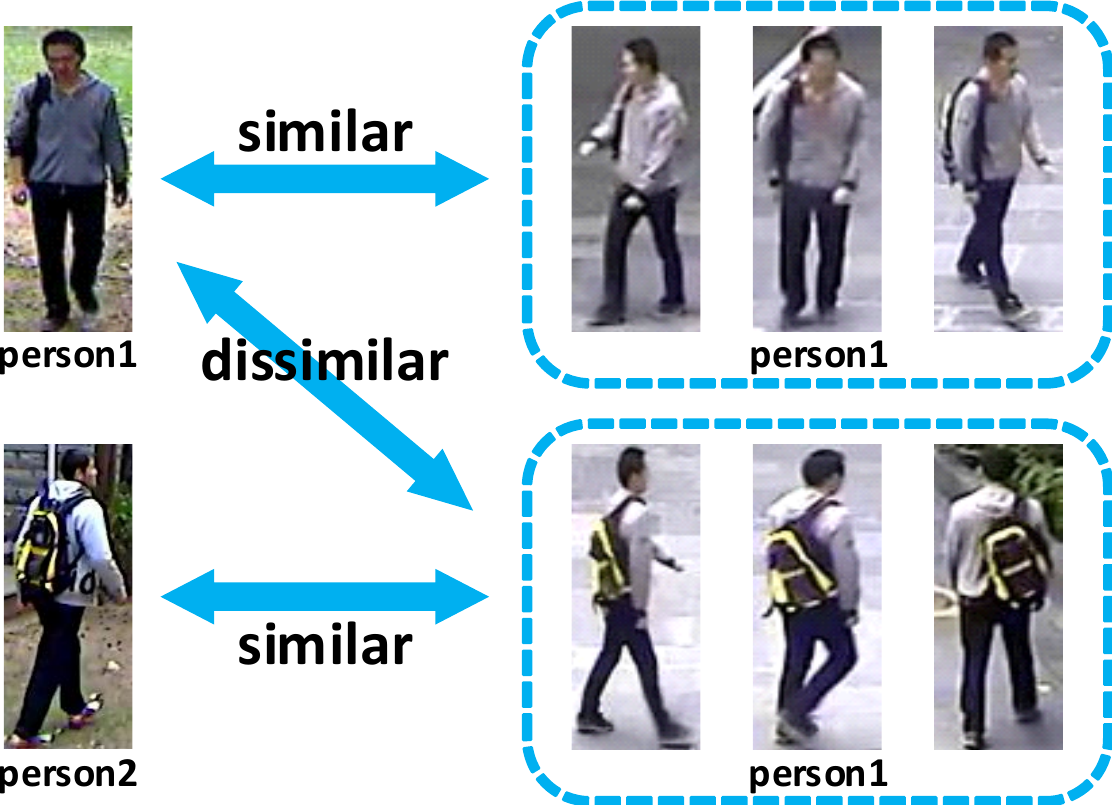}
  \caption{Person appearances are similar in the same orientation, but dissimilar in different orientation. \textbf{Best viewed in color.}}\label{fig_dissimilar}
\end{figure}

During matching stage, for each frame existing in the probe bag, ODBoA selects the most suitable frame from the gallery bag using orientation information. We design a selection strategy based on the following observations of orientation misalignment as illustrated in Fig. \ref{fig_dissimilar}: (1) Appearances obtained in different orientations of one person may vary significantly, which results in large intra-class dissimilarity; (2) Appearances obtained in some orientations of different persons may be similar to some extend, which results in large inter-class ambiguity; (3) If integrating the appearances into a single signature directly in feature level fusion, or using average or minimal distance in decision level fusion, the information from other orientations may become some kind of noise for person matching.
Therefore, orientation information plays an important role in integrating multi-shot appearances. 

\begin{figure}[t]
  \centering
  \includegraphics[width=8.5cm]{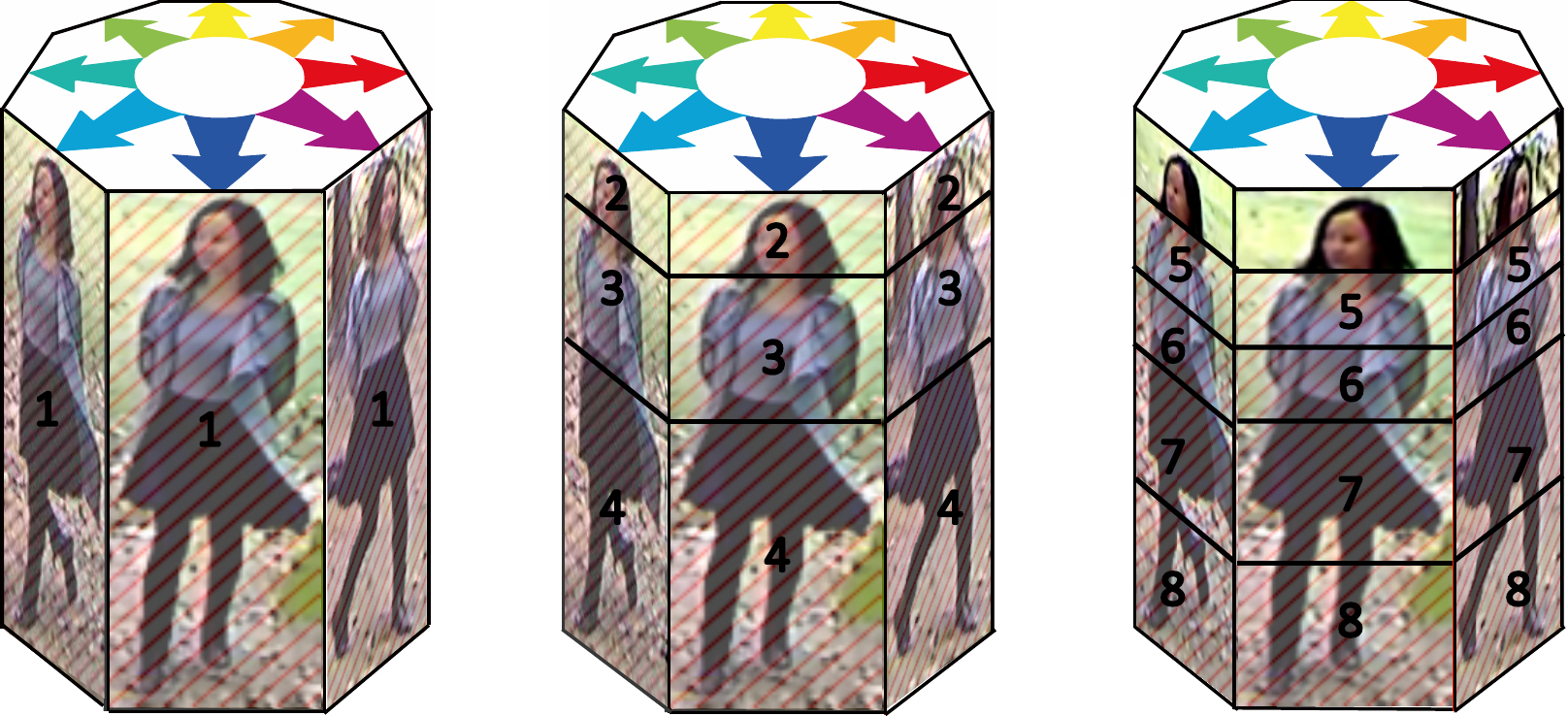}
  \centerline{(a)~~~~~~~~~~~~~~~~~~~~~~(b)~~~~~~~~~~~~~~~~~~~~~~(c)}\medskip
  \caption{The orientation driven body-structure pyramid. Each level contains eight orientation with different spatial partitioning as shown in Fig. \ref{fig_pyramid}(a)(b)(c). \textbf{Best viewed in color.}}\label{fig_360pyramid}
\end{figure}

As illustrated in Fig. \ref{fig_ODBoA}, the selection is based on person orientation information. To be specific, the frame with the same orientation is selected if gallery bag contains it, or else the frame with the adjacent orientation is selected. Then, if no frame with the same or adjacent orientation exists, we randomly select one frame from gallery bag.
After selecting all suitable frames, we employed an effective pooling strategy which is guided by the orientation driven body-structure pyramid as shown in Fig. \ref{fig_360pyramid}. It is an extension of body-structure pyramid mentioned in Sec. \Rmnum{3}, $i.e.$, pooling the feature vectors in the same part obtained from multiple frames into a single vector.
Considering that person appearances obtained in different orientation may vary significantly, we generate multi-shot signature by using max pooling which can capture the salient difference over sparse codes well \cite{ScSPM}.
Traditional methods always fuse appearance information from multiple frames in either low feature level or decision level. However, we extract the appearance information based on BSFR and fuse multi-shot appearances with mid-level feature fusion. The mid-level feature fusion incorporates person body structure information and captures salient properties of local regions via max pooling, which is more suitable for the multi-shot person image fusion problem.
Similarity calculation between the probe and gallery signature will be introduced in the next subsection.
The proposed matching strategy is very helpful to solve data imbalanced problem which may introduce much dissimilar appearance noise as shown in Fig. \ref{fig_dissimilar}.

Algorithm 1 shows the self-explanatory pseudo code of ODBoA matching strategy. First, corresponding frames are selected for each person using our selection strategy. Then, the max pooling is employed to integrate these frames into a single signature. Finally, the similarity between two signatures is calculated.

\floatname{algorithm}{Algorithm}
\renewcommand{\algorithmicrequire}{\textbf{Input:}}
\renewcommand{\algorithmicensure}{\textbf{Output:}}
\begin{algorithm}[t]
    \caption{ODBoA Matching}
    \begin{algorithmic}[1] 
        \Require Probe ODBoA model $B_{p}=\{b_{p1},b_{p2},..,b_{p8}\}$; \ Gallery ODBoA model $B_{g}=\{b_{g1},b_{g2},..,b_{g8}\}$
        \Ensure Similarity score $S$
        \State \% Selection Strategy
        \State $B_{pSel} \gets \{\}$ \% store the elements selected from $B_{p}$
        \State $B_{gSel} \gets \{\}$ \% store the elements selected from $B_{g}$
        \For{$i = 1 \to 8$}
            \If{$b_{pi}$ \ is \ not \ empty} \% $i$-th orientation exists
                \If{$b_{gi}$ \ is \ not \ empty} \% select same orientation
                    \State $B_{pSel} \gets B_{pSel} \cup b_{pi}$
                    \State $B_{gSel} \gets B_{gSel} \cup b_{gi}$
                \Else \ \% select adjacent orientation
                    \State $b_{adj} \gets Adjacent(b_{gi})$
                    \If{$b_{adj} \ is \ not \ empty$}
                        \State $B_{pSel} \gets B_{pSel} \cup b_{pi}$
                        \State $B_{gSel} \gets B_{gSel} \cup b_{adj}$
                    \EndIf
                \EndIf
            \EndIf
        \EndFor
        \If{$B_{gSel}$ \ is \ empty}
            \State $Q \gets min($\textbf{NUM}$(B_{p}), $\textbf{NUM}$(B_{g}))$
            \State random select $Q$ elements from $B_{p}$ for $B_{pSel}$
            \State random select $Q$ elements from $B_{g}$ for $B_{gSel}$
        \EndIf
        \State $f_{p} \gets$ MaxPooling $B_{pSel}$ ~~\% Pooling
        \State $f_{g} \gets$ MaxPooling $B_{gSel}$ ~~\% Pooling
        \State $S \gets$ Score($f_{p},f_{g}$) ~~\% Similarity Score Calculation
        \State return $S$
        \State \% Calculate the number of valid orientations in ODBoA
        \Function{\textbf{NUM}}{ODBoA $B=\{b_{1},b_{2},..,b_{8}\}$}
            \State $N \gets$ $0$
            \For{$i = 1 \to 8$}
                \If{$b_{i}$ \ is not \ empty}
                    \State $N \gets$ $N+1$
                \EndIf
            \EndFor
            \State return $N$
        \EndFunction
    \end{algorithmic}
\end{algorithm}

\begin{figure*}[htb]
  \centering
  \includegraphics[width=18cm]{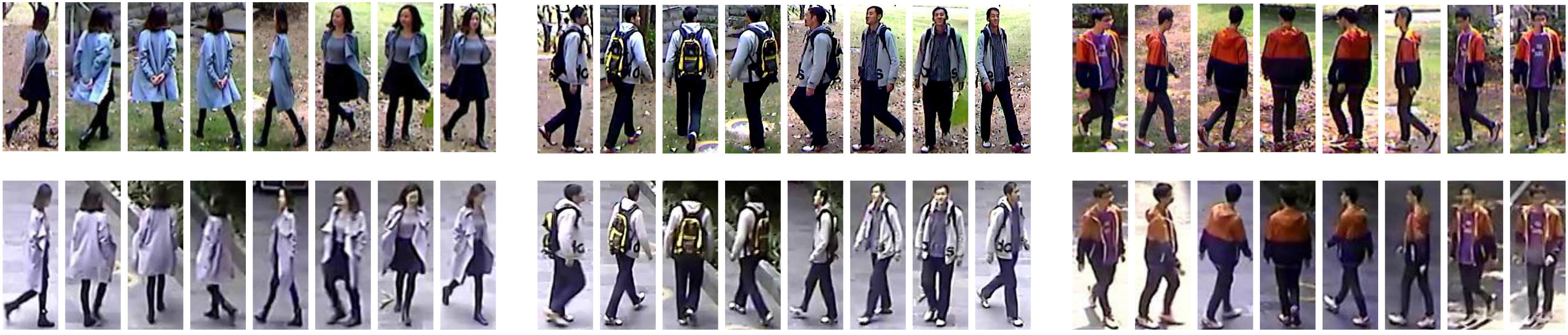}
  \caption{Sample images of the PKU-Reid dataset. All images are normalized to 128 $\times$ 48 (Top:) Sample images of three identities with distinctive appearance captured by camera A. (Bottom:) Sample images of the same three identities  captured by camera B.}\label{fig_PKU-Reid}
\end{figure*}

\subsection{KISSME-based similarity calculation}
Three low-level descriptors are used in our method, including:
1) weighted HSV (wHSV) color histograms are extracted to capture color information as suggested in \cite{SDALF}; 2) dense SIFT descriptors are used to capture texture information and handle illumination variation; 3) LAB color histograms are extracted to enhance illumination invariance. The encoded wHSV, LAB, SIFT feature vectors are denoted as $wH(I)$, $LAB(I)$, $SIFT(I)$ respectively, and $I$ is the pedestrian image. 

Here, Mahalanobis distance is used to measure the distance between feature vector $x_i$ and $x_j$ as follows,
\begin{eqnarray}\label{eq8}
    d_M^2(x_i,x_j) = (x_i-x_j)^\top M (x_i-x_j)
\end{eqnarray}
where $M$ denotes the metric matrix. In order to process large-scale person re-identification data, KISSME \cite{KISSME} is applied to learn the Mahalanobis distance metric.
KISSME is established at a statistical inference point of view that the optimal statistical decision whether a pair $(i,j)$ is dissimilar or not can be obtained by a likelihood ratio test as follows,
\begin{equation}\label{eq9}
    \begin{aligned}
    \delta(x_{ij})=log\begin{pmatrix}\cfrac{p(x_{ij}|H_0)}{p(x_ij|H_1)}\end{pmatrix}=log\begin{pmatrix}\cfrac{f(x_{ij}|\theta_0)}{f(x_ij|\theta_1)}\end{pmatrix} \\
    \end{aligned}
\end{equation}
where $x_{ij}=x_i-x_j$ denotes the pairwise difference with zero mean.
$H_0$ and $H_1$ denote the hypothesises that a pair is dissimilar and a pair is similar, respectively.
$f(x_{ij}|\theta_0)$ and $f(x_{ij}|\theta_1)$ are the corresponding probability distribution functions with parameters $\theta_0$, $\theta_1$.
A high value of $\delta(x_{ij})$ means that $H_0$ is validated and pair $(i,j)$ is dissimilar. In contrast, a low value means that $H_1$ is rejected and pair $(i,j)$ is similar.
In order to simplify the problem, the data are assumed to obey gaussian distribution,
\begin{equation}\label{eq10}
    \begin{aligned}
    \delta(x_{ij})&=log\begin{pmatrix}\cfrac{\cfrac{1}{\sqrt{2\pi|\sum_{y_{ij}=0}|}} exp(-1/2x_{ij}^T\sum_{y_{ij}=0}^{-1}x_{ij})}{\cfrac{1}{\sqrt{2\pi|\sum_{y_{ij}=1}|}} exp(-1/2x_{ij}^T\sum_{y_{ij}=1}^{-1}x_{ij})}\end{pmatrix} \\
    &= x_{ij}^T(\sum\nolimits_{y_{ij}=0}^{-1}-\sum\nolimits_{y_{ij}=1}^{-1})x_{ij}+C \\
    \end{aligned}
\end{equation}
where $C=log(|\sum\nolimits_{y_{ij}=1}|)-log(|\sum\nolimits_{y_{ij}=0}|)$ is a constant term, which just provides an offset and can be ignored here. Therefore, the metric matrix can be calculated as follows,
\begin{equation}\label{eq11}
    \begin{aligned}
    M=(\sum\nolimits_{y_{ij}=0}^{-1}-\sum\nolimits_{y_{ij}=1}^{-1}) .
    \end{aligned}
\end{equation}
However, the feature vectors can not be processed by KISSME directly since their high-dimension may result in a singular matrix during metric learning. Therefore, the feature vector is reduced to a low-dimension space as most existing literatures \cite{KISSME,SCNCD}.

\subsection{Distance fusion}
Since different kinds of feature show different discrimination, a decision level fusion strategy is used to integrate the contributions of different features as follows,
\begin{equation}\label{eq12}
    \begin{aligned}
    d&(I_A,I_B) = \beta_{wH} \cdot d_{wH}(wH(I_A),wH(I_B)) \\
        &+ \beta_{LAB} \cdot d_{LAB}(LAB(I_A),LAB(I_B)) \\
        &+ \beta_{SIFT} \cdot d_{SIFT}(SIFT(I_A),SIFT(I_B))
    \end{aligned}
\end{equation}
where $d_{wH}$, $d_{LAB}$, and $d_{SIFT}$ are the normalized feature vector distances
calculated by Eq. \eqref{eq8}, and $\beta_{wH}$, $\beta_{LAB}$, $\beta_{SIFT}$ denote the corresponding integrating weights.

\begin{figure*}[htb]
\begin{minipage}[b]{.31\linewidth}
  \centering
  \centerline{\includegraphics[width=6cm]{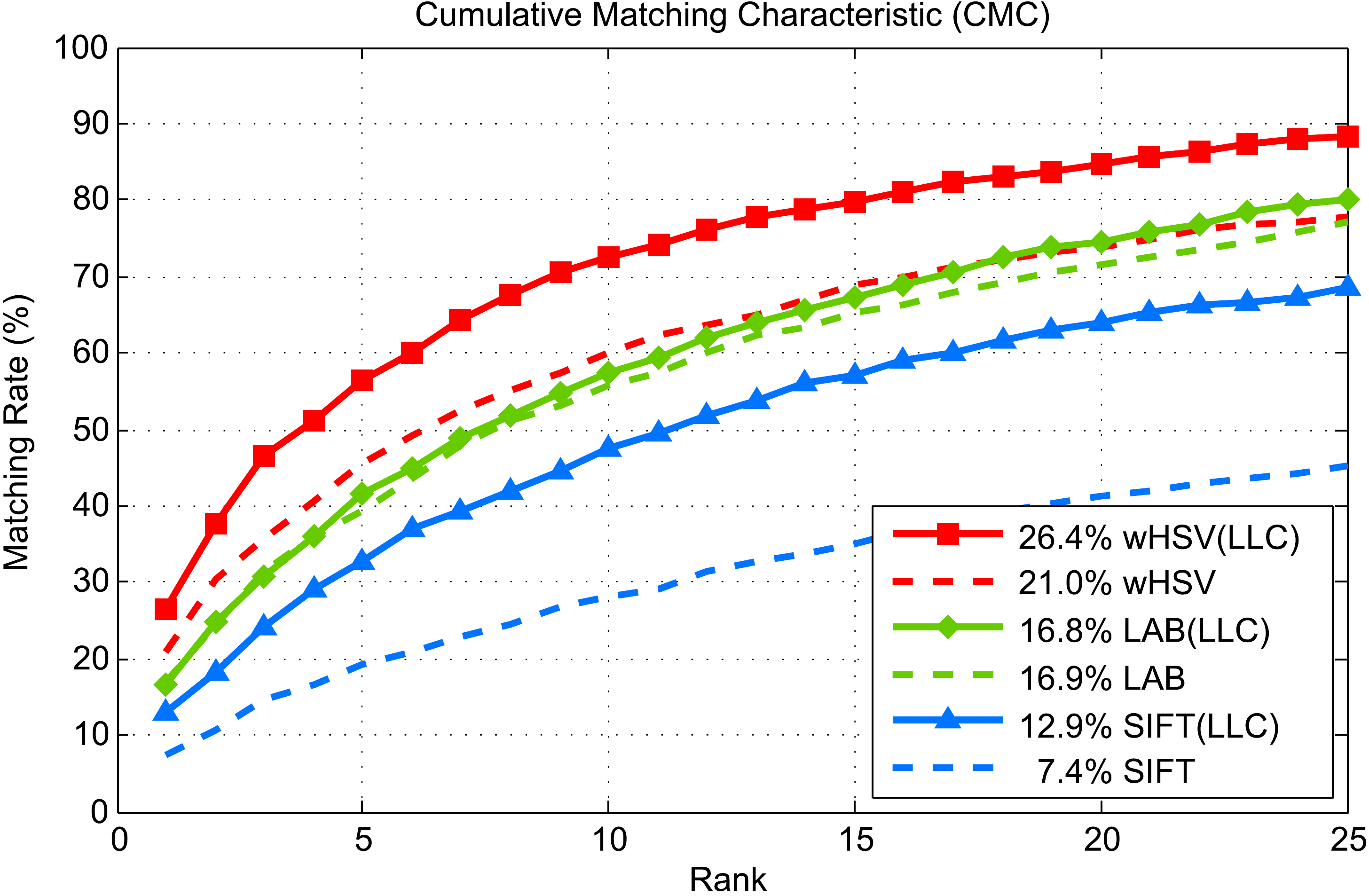}}
  \centerline{(a) Evaluation of LLC coding}\medskip
\end{minipage}
\hfill
\begin{minipage}[b]{.31\linewidth}
  \centering
  \centerline{\includegraphics[width=6cm]{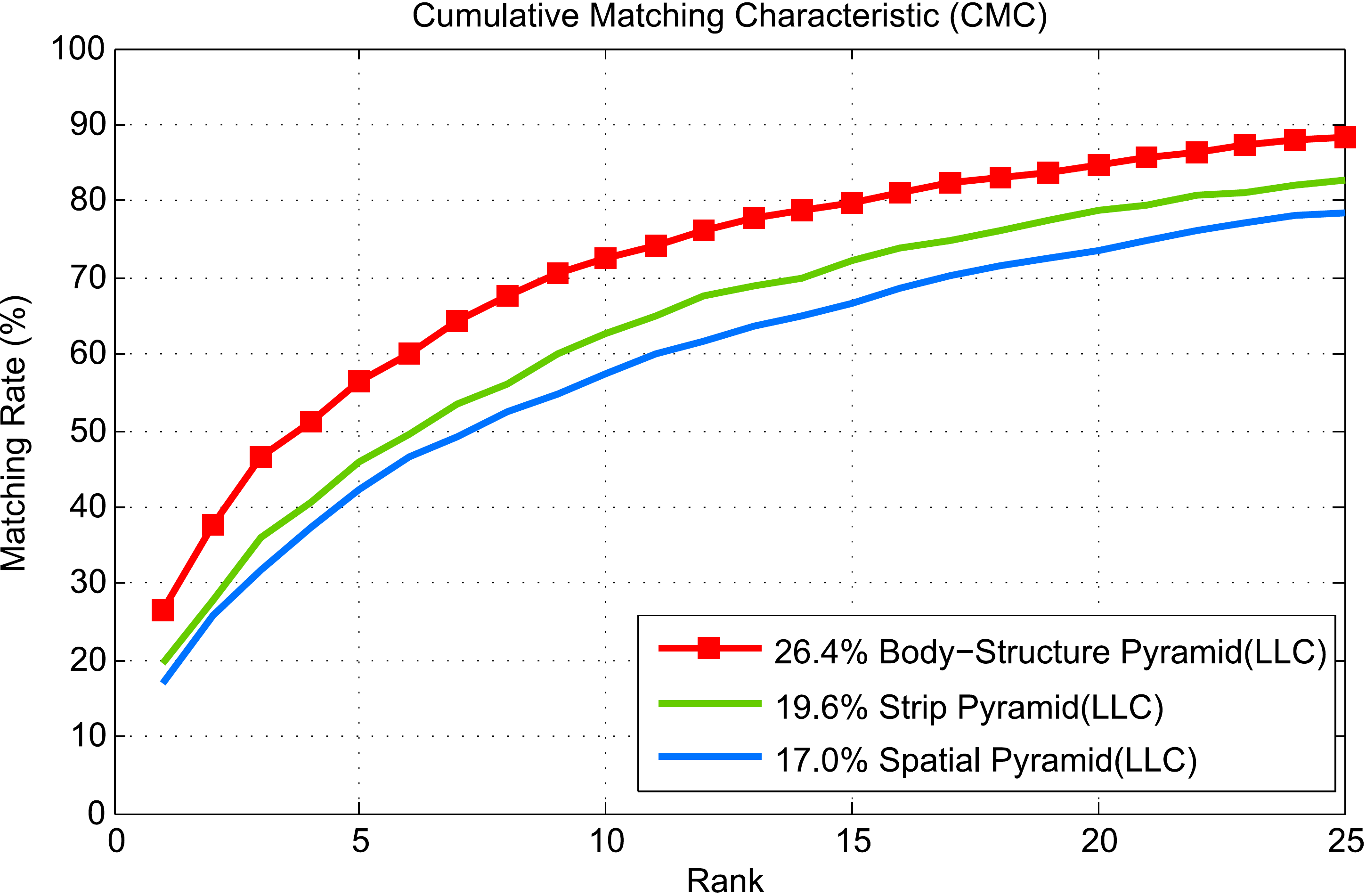}}
  \centerline{(b) Evaluation of body-structure pooling}\medskip
\end{minipage}
\hfill
\begin{minipage}[b]{0.31\linewidth}
  \centering
  \centerline{\includegraphics[width=6cm]{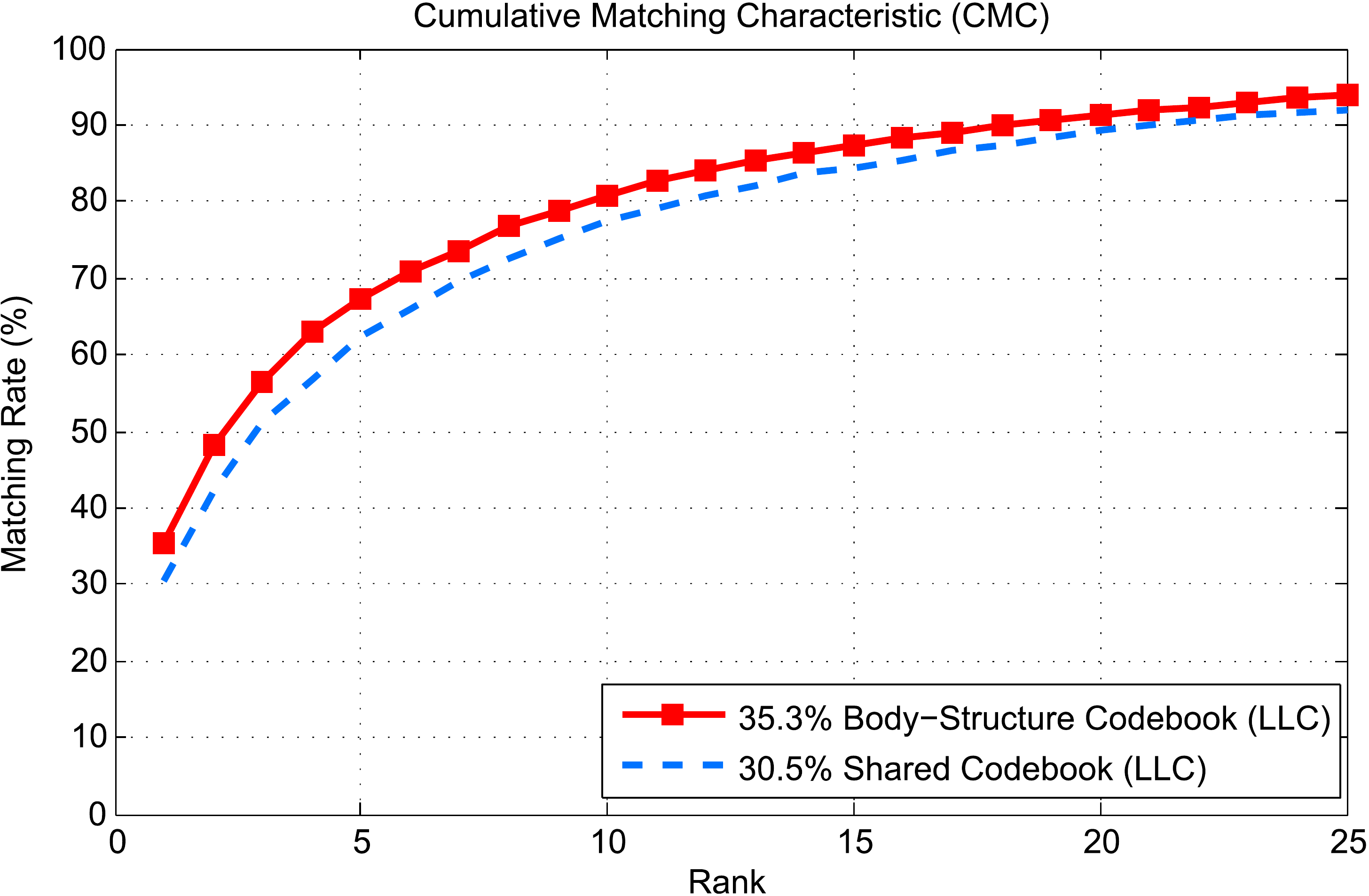}}
  \centerline{(c) Evaluation of body-structure codebook}\medskip
\end{minipage}
\caption{Evaluations on the VIPeR dataset. Rank-1 matching rate is marked before the name of each approach. \textbf{Best viewed in color.}}\label{fig_evaOnViper}
\end{figure*}

\section{Datasets and evaluation protocol}
Our approach is evaluated on three publicly challenging datasets, VIPeR \cite{VIPeR}, CUHK01 \cite{CUHK}, TUD \cite{CVPR10_TUD} and two newly proposed datasets PKU-Reid, Market-1203.

\textbf{VIPeR dataset$^1$} \footnotetext[1]{\url{http://vision.soe.ucsc.edu/?q=node/178}} contains 632 pedestrian image pairs
taken from arbitrary viewpoints under varying illumination conditions. All images are normalized to 128 $\times$ 48 pixels. This dataset is randomly split into two parts, both consisting of 316 individuals, one for training and the other for testing. 

\textbf{CUHK01 dataset$^2$} \footnotetext[2]{\url{http://www.ee.cuhk.edu.hk/~xgwang/CUHK_identification.html}}contains 971 individuals captured from two disjoint camera views. Under each camera view, one person has two images which are normalized to 160 $\times$ 60 pixels. This dataset is split into two parts randomly. One contains 485 individuals for training, and the other contains 486 individuals for testing. As each person has two images in probe and gallery, respectively, the four distances between image pairs are averaged to obtain the final distance following \cite{CUHK_MidLevel}. 

\textbf{PKU-Reid dataset$^4$}
\footnotetext[4]{\url{https://github.com/charliememory/PKU-Reid-Dataset.git}}contains 114 individuals including 1824 images captured from two disjoint camera views. For each person, eight images are captured from eight different orientations under one camera view and are normalized to 128 $\times$ 48 pixels. This dataset is also split into two parts randomly. One contains 57 individuals for training, and the other contains 57 individuals for testing. 
To the best of our knowledge, PKU-Reid dataset is the first one that collects person appearances in all eight orientations.

\textbf{Market-1203 dataset$^5$}
\footnotetext[5]{\url{https://github.com/charliememory/Market1203-Reid-Dataset.git}}contains 1203 individuals captured from two disjoint camera views. For each person, one to twelve images are captured from one to six different orientations under one camera view and are normalized to 128 $\times$ 64 pixels.
This dataset is constructed based on the Market-1501 benchmark data and we annotate the orientation label for each image manually. 
We randomly select 601 individuals for training and the rest for testing.

\textbf{3DPeS dataset$^6$}
\footnotetext[6]{\url{http://www.openvisor.org/3dpes.asp}}contains different sequences of 200 individuals taken from eight static disjoint cameras in an outdoor scenario. Strong variations in viewpoints and lighting conditions make this dataset very challenging for person re-identification. In order to compare the results of our method with previous works, we use the same setup as \cite{ICPR14_Ori_Re-id}, that is only 190 people are randomly chosen, half for training and half for testing.

\section{Experiments and discussions}
The detailed parameters are set as follows:
images are divided into overlapping patches of size $8 \times 8$ with $4 \times 4$ stride.
Body-structure codebook contains eight sub-codebooks corresponding to eight body parts as shown in Fig. \ref{fig_pyramid}.
Each sub-codebook containing 1024 entities is constructed with 5000 patches randomly selected from the corresponding body part patch set.
$\beta_{wH}=2$, $\beta_{LAB}=1$, $\beta_{SIFT}=1$ is set for VIPeR, PKU-Reid, Market-1203, 3DPeS datasets empirically. $\beta_{wH}=1$, $\beta_{LAB}=1$, $\beta_{SIFT}=1$ is set for CUHK01 dataset, since higher image resolution may lead to more reliable SIFT descriptors.
Experimental results are reported in the form of average Cumulated Matching Characteristic (CMC) curve for 10 trials.

\begin{figure}[t]
\hspace{3mm}
\begin{minipage}[b]{.40\linewidth}
  \centering
  \centerline{\includegraphics[width=4.1cm]{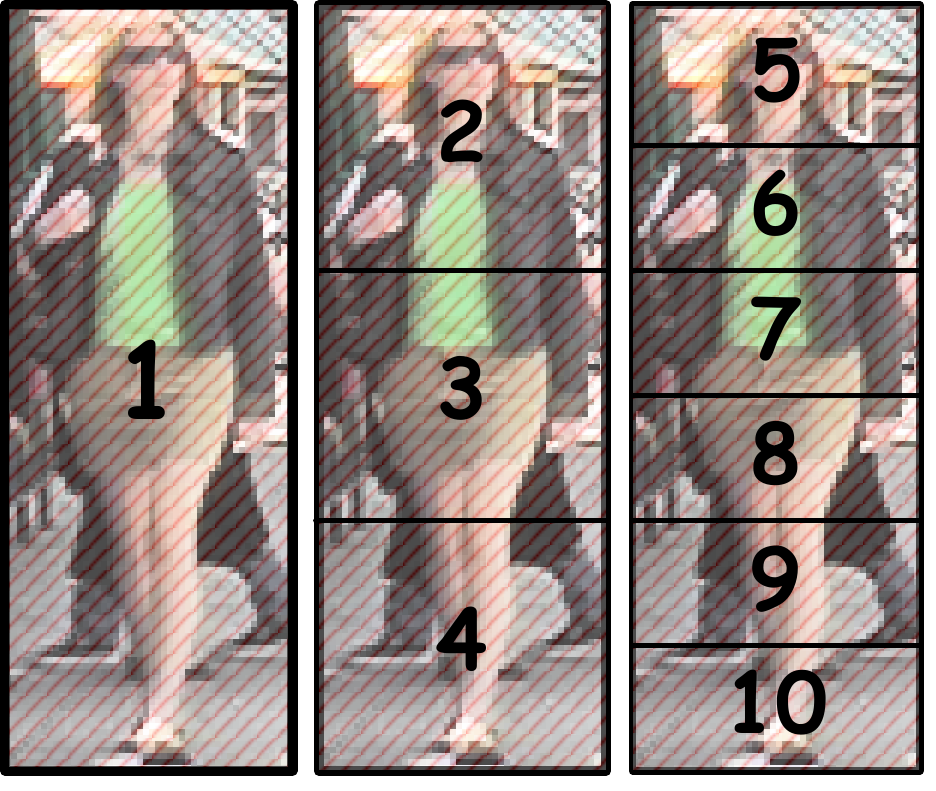}}
  \centerline{(a) Strip pyramid}\medskip
\end{minipage}
\vspace{-2.5mm}
\hspace{7mm}
\begin{minipage}[b]{.40\linewidth}
  \centering
  \centerline{\includegraphics[width=4.1cm]{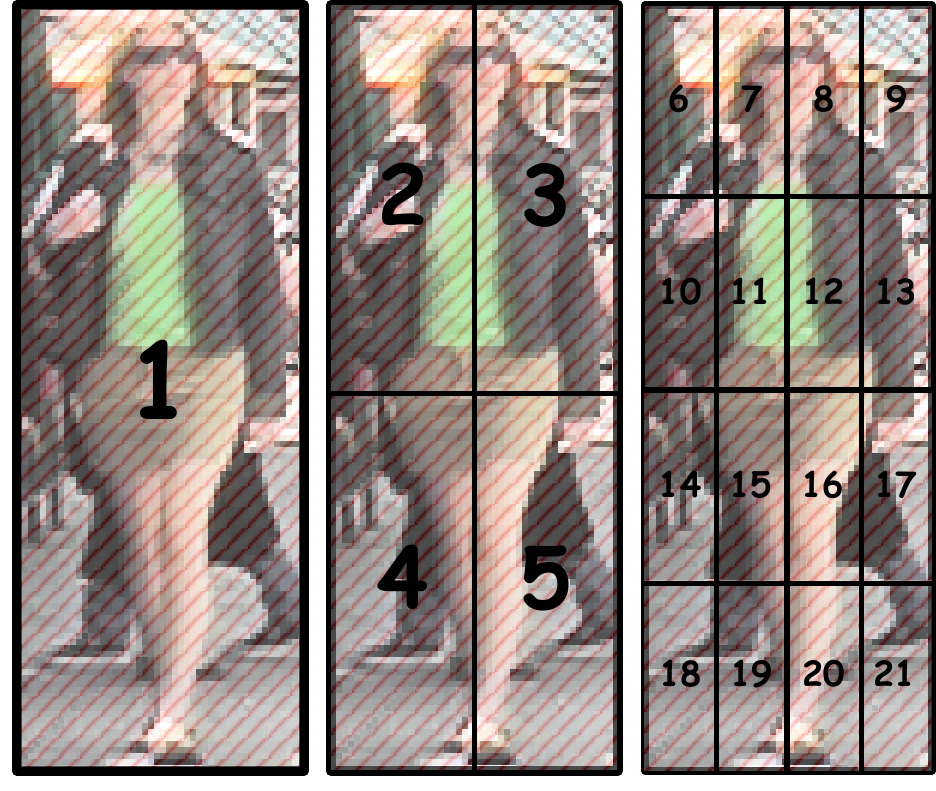}}
  \centerline{(b) Spatial pyramid}\medskip
\end{minipage}
\caption{Two compared pyramid structures. (a) Strip pyramid contains three layers which divide body into one, three, six equal horizontal strips, respectively. (b) Spatial pyramid contains three layers which divide body into 1 $\times$ 1, 2 $\times$ 2, 4 $\times$ 4 equal subdivisions, respectively.}\label{fig_twoPyramid}
\end{figure}

\subsection{Performances of Person Feature Representation}
The effectiveness of LLC coding strategy, body-structure pooling and body-structure codebook are all evaluated on VIPeR. Comparisons between BSFR and state-of-the-arts \cite{KISSME,SDALF,BiCov,CUHK_SalienceMatching,AttributeTopicModel,CUHK_MidLevel,SCNCD} are done on VIPeR and CUHK01 datasets.
The dimension of the feature vectors is reduced to 74 with gauss kernel PCA empirically for both datasets. The gauss kernel band width is set to 0.6 and 0.8 for VIPeR and CUHK01, respectively.

\textbf{Evaluation of LLC.}
LLC encodes low-level descriptors into mid-level features. Evaluation of LLC compares the performances using three low-level descriptors with and without LLC coding using shared codebook ($i.e.$, the commonly used codebook in LLC \cite{LLC}).
All features are pooled via body-structure pooling. Fig. \ref{fig_evaOnViper}(a) shows that for all the three low-level descriptors, performances using LLC are more competitive than that without using LLC, especially for wHSV and SIFT.
Taking rank 10 for example, an improvement of 12.4\% for wHSV and 19.3\% for SIFT are respectively achieved. 
The main reason is that as an extension of BoF, LLC is good at handling space misalignment caused by different viewpoints and poses. Furthermore, the locality property of LLC can generate similar codes for similar descriptors, which may improve the feature discrimination.

\begin{table}[t]
\centering
\caption{Comparisons with the state-of-the-arts on VIPeR}\label{table1}  

\begin{tabular*}{8.5cm}
{@{\extracolsep{\fill}} c c c c c}
\toprule[2pt]
VIPeR & Rank 1 & Rank 10 & Rank 20 & Rank 50 \\
\midrule[1pt]
ELF\cite{ELF} & 12 & 43 & 60 & 81 \\
EIML\cite{EIML} & 22 & 63 & 78 & 93 \\
KISSME\cite{KISSME} & 19.6 & 62.2 & 77 & 91.8\\
SDALF\cite{SDALF} & 19.9 & 49.4 & 65.7 & 84.8 \\
eBiCov\cite{BiCov} & 20.7 & 56.2 & 68.0 & - \\
Salience\cite{CUHK_SalienceMatching} & 30.2 & 65.5 & 79.2 & - \\
ARLTM\cite{AttributeTopicModel} & 21.2 & 38.7 & 52.9 & 67.5 \\
Mid-Filters\cite{CUHK_MidLevel} & 29.1 & 65.6 & 79.9 & - \\
RD\cite{TCSVT15_reid_RD} & 33.3 & 78.4 & 88.5 & 97.5 \\
SCNCD$_{all}$\cite{SCNCD} & 33.7 & 74.8 & 85.0 & 93.8 \\
\midrule[1pt]
\textbf{BSFR}(Ours) & \textbf{35.3} & \textbf{80.8} & \textbf{91.2} & \textbf{98.4} \\
\bottomrule[2pt]
\end{tabular*}

\end{table}

\begin{table}[t]
\centering
\caption{Comparisons with the state-of-the-arts on CUHK01}\label{table2}  

\begin{tabular*}{8.5cm}
{@{\extracolsep{\fill}} c c c c c}
\toprule[2pt]
CUHK01 & Rank 1 & Rank 10 & Rank 20 & Rank 50 \\
\midrule[1pt]
SDALF\cite{SDALF} & 9.9 & 30.3 & 41.0 & - \\
ITML\cite{CUHK_MidLevel} & 16.0 & 45.6 & 59.8 & - \\
GenericMetric\cite{CUHK} & 20.0 & 50.0 & 69.3 & - \\
Salience\cite{CUHK_SalienceMatching} & 28.5 & 55.7 & 68.0 & - \\
RD\cite{TCSVT15_reid_RD} & 31.1 & 68.6 & 79.2 & 90.4 \\
Mid-Filters\cite{CUHK_MidLevel} & 34.3 & 65.0 & 75.0 & - \\
\midrule[1pt]
\textbf{BSFR}(Ours) & \textbf{37.4} & \textbf{73.3} & \textbf{84.1} & \textbf{93.5} \\
\bottomrule[2pt]
\end{tabular*}
\end{table}

\textbf{Evaluation of body-structure pooling.} To validate the effectiveness of body-structure pyramid on feature pooling, we compare the matching results of wHSV features pooled by three different spatial structures: body-structure pyramid, strip pyramid and spatial pyramid.
Fig. \ref{fig_twoPyramid} depicts the detailed structures of strip pyramid and spatial pyramid.
It is noted that traditional person re-identification methods usually divided person image into several equal horizontal strips. For fair comparison, strips pyramid is constructed as shown in Fig. \ref{fig_twoPyramid}(a), which has three levels of spatial partitioning as 1 $\times$ 1, 3 $\times$ 1, 6 $\times$ 1 with totally 1+3+6=10 spatial cells. While spatial pyramid \cite{SPM} is a classical spatial structure as shown in Fig. \ref{fig_twoPyramid}(b), which also has three levels of spatial partitioning as 1 $\times$ 1, 2 $\times$ 2, 4 $\times$ 4 with totally 1+4+16=21 spatial cells.
All features are encoded by LLC using shared codebook.
Fig. \ref{fig_twoPyramid}(b) shows that our proposed feature pooling guided by body-structure pyramid produces a remarkable performance improvement over strip pyramid and spatial pyramid across a large range of ranks.
The reasonable explanation is that our proposed body-structure pyramid accords with human body structure better.
Furthermore, feature pooling is a good way to integrate the body-structure information into feature representation.

\begin{figure}[t]
  \centering
  \includegraphics[width=8.5cm]{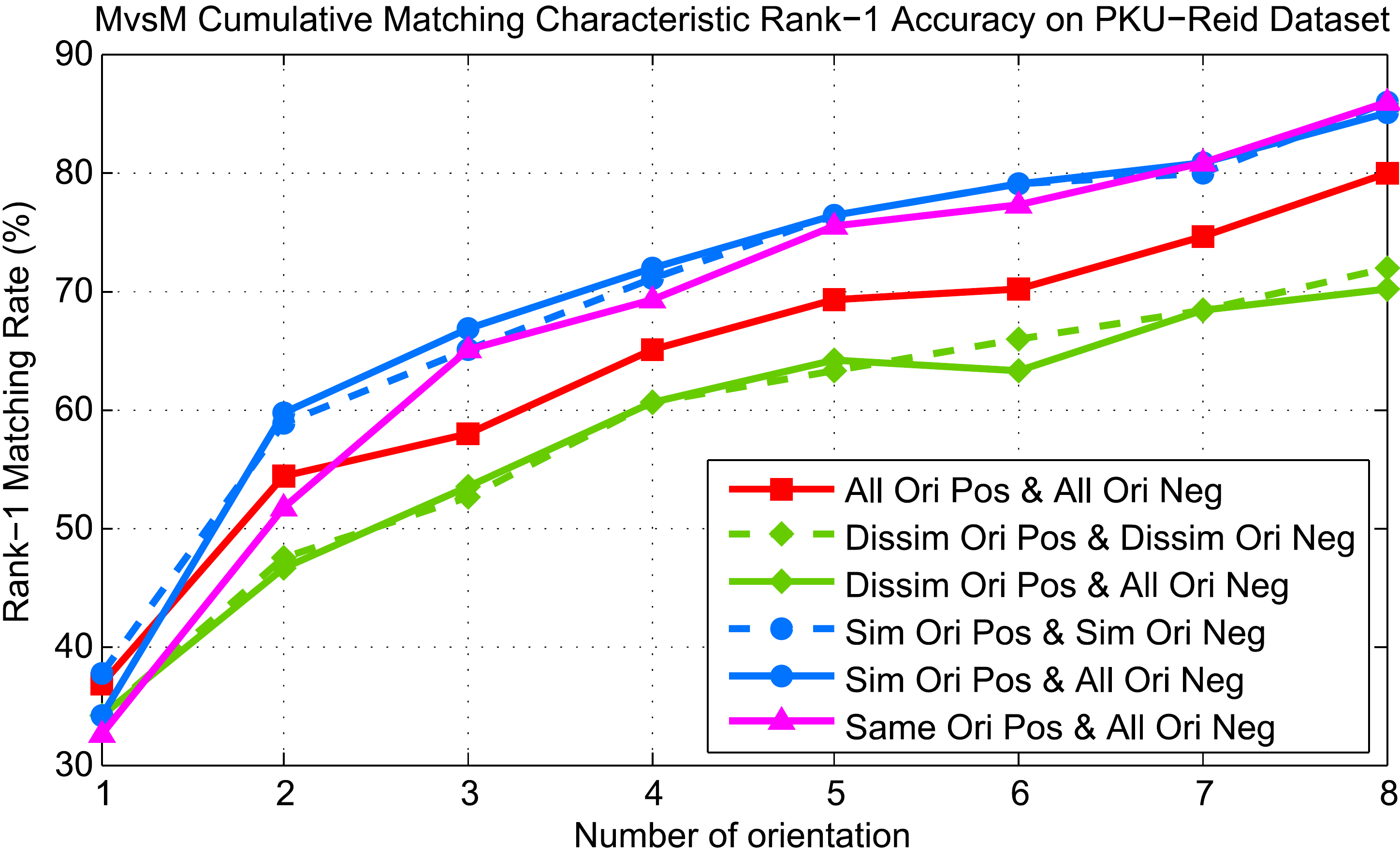}
  \vspace{-3mm}
  \caption{Helpfulness of person orientation information on metric model training. Performances are evaluated on PKU-Reid dataset with ODBoA-Avg method. \textbf{Best viewed in color.}}\label{fig_oriEva}
\end{figure}

\begin{figure}[t]
\hspace{0.5mm}
\begin{minipage}[b]{.43\linewidth}
  \centering
  \centerline{\includegraphics[width=4.1cm]{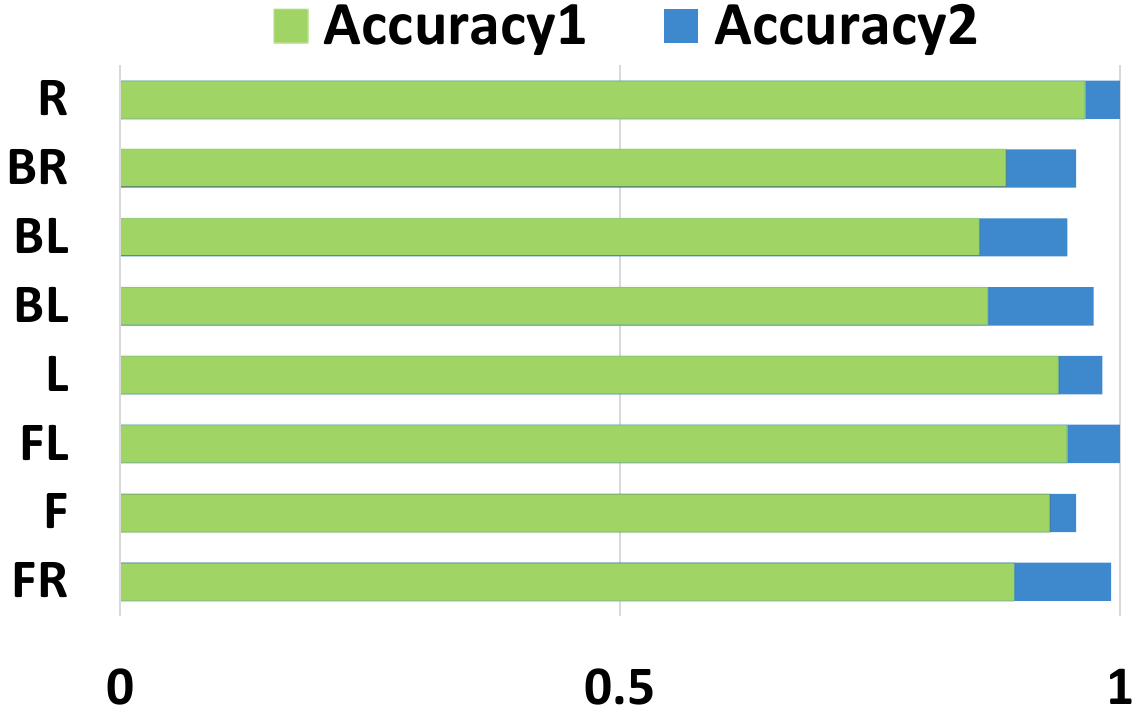}}
  \centerline{(a) Evaluation on PKU-Reid}\medskip
\end{minipage}
\vspace{-2mm}
\hspace{7mm}
\begin{minipage}[b]{.43\linewidth}
  \centering
  \centerline{\includegraphics[width=4.1cm]{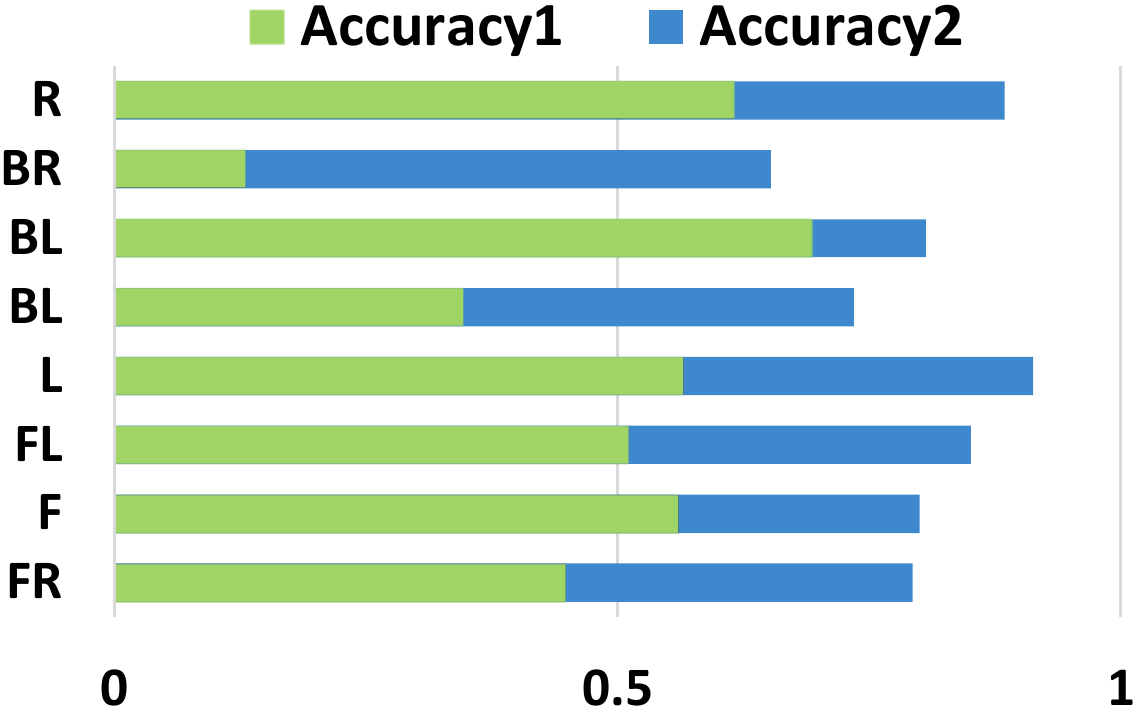}}
  \centerline{(b) Evaluation on TUD}\medskip
\end{minipage}
\vspace{-2mm}
\caption{Evaluations of person orientation estimation. Eight quantized orientations are considered: Right (R), Back-Right (BR), Back (B), Back-Left (BL), Left (L), Front-Left (FL), Front (F), Front-Right (FR). Accuracy1: result is correct when the predicted and true orientation are same. Accuracy2: result is correct when the predicted and true orientation are same or adjacent. \textbf{Best viewed in color.}}\label{fig_oriEstimation}
\vspace{-2.5mm}
\end{figure}

\textbf{Evaluation of body-structure codebook.}
To evaluate the effectiveness of body-structure codebook, wHSV, LAB and SIFT features encoded by LLC with body-structure pooling are employed.
As depicted in Fig. \ref{fig_evaOnViper}(c), body-structure codebook achieves better performance than shared codebook,
since it can reflect characteristics of different body parts more accurately.

\textbf{Comparison with state-of-the-arts.}
Comparing experiments of our BSFR and the state-of-the-art methods are conducted on VIPeR and CUHK01 datasets.
Table \ref{table1} and Table \ref{table2} show that BSFR outperforms other state-of-the-art methods on both datasets.
The reasonable explanation is that BSFR makes full use of body structure information and uses mid-level features coded by LLC, which are insensitive to space misalignment and robust to the variations of pose and viewpoint. In addition, our mid-level features are encoded using LLC with better discrimination and low computation complexity linear to the size of codebook and the number of the sampled patches.

\begin{figure}[t]
  \centering
  \includegraphics[width=8.5cm]{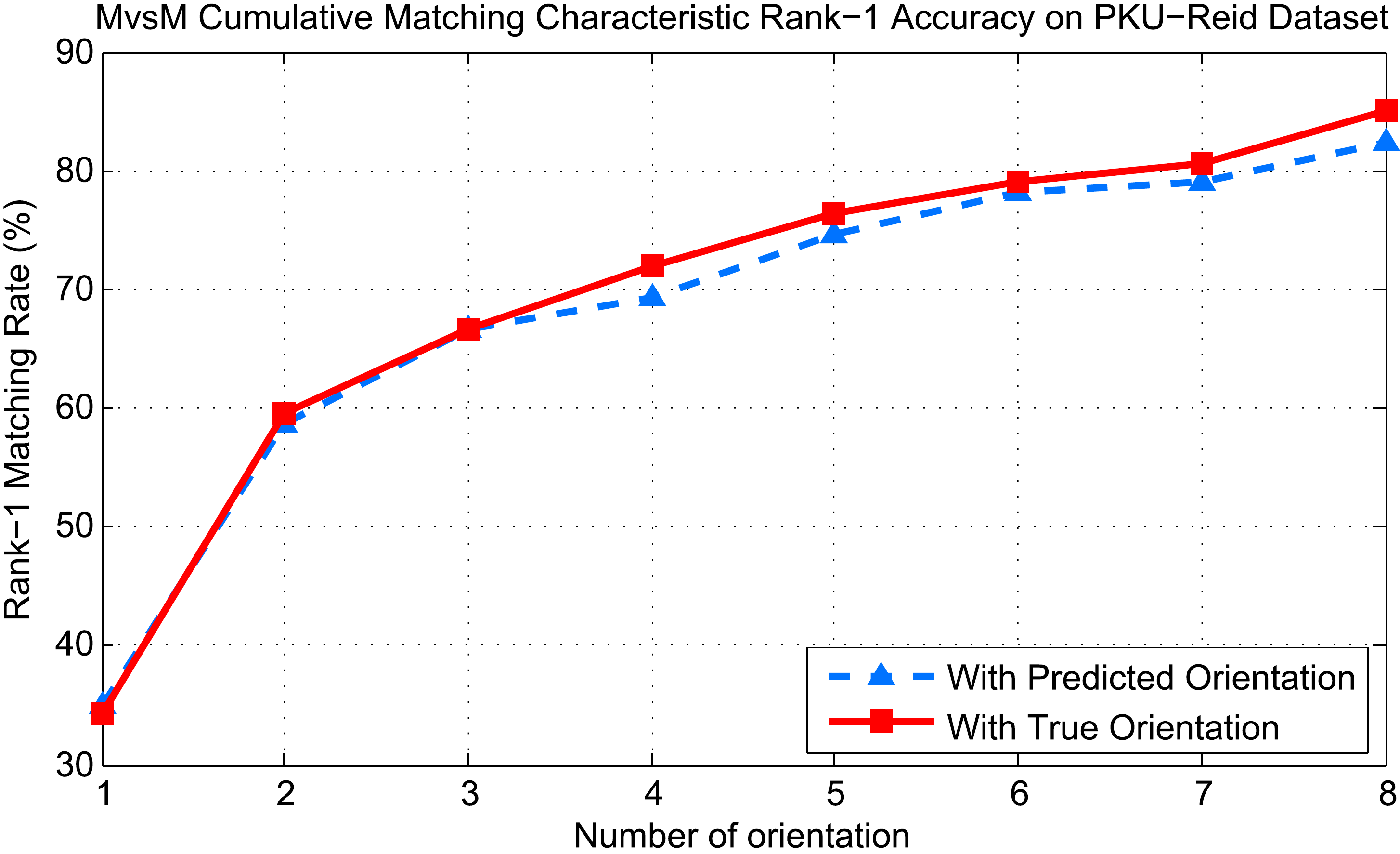}
  \vspace{-3mm}
  \caption{Influence of orientation estimation accuracy on person re-identification. Performances are evaluated on PKU-Reid dataset with ODBoA-Avg method.}\label{fig_ori4reid}
  \vspace{-3mm}
\end{figure}


\begin{figure*}[t]
\begin{minipage}[b]{.31\linewidth}
  \centering
  \centerline{\includegraphics[width=6cm]{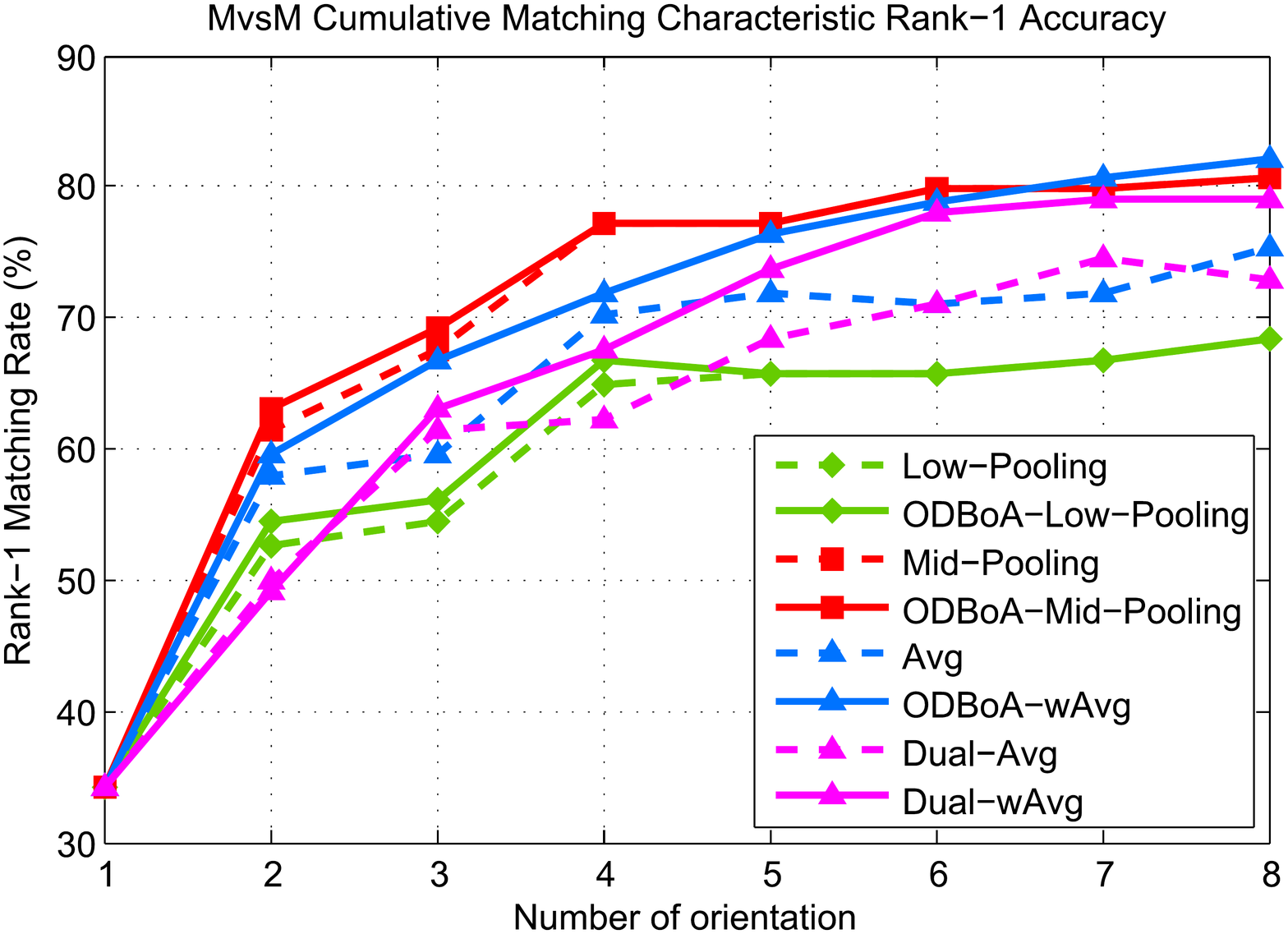}}
  \centerline{(a) Evaluations on PKU-Reid}
\end{minipage}
\hfill
\begin{minipage}[b]{.31\linewidth}
  \centering
  \centerline{\includegraphics[width=6cm]{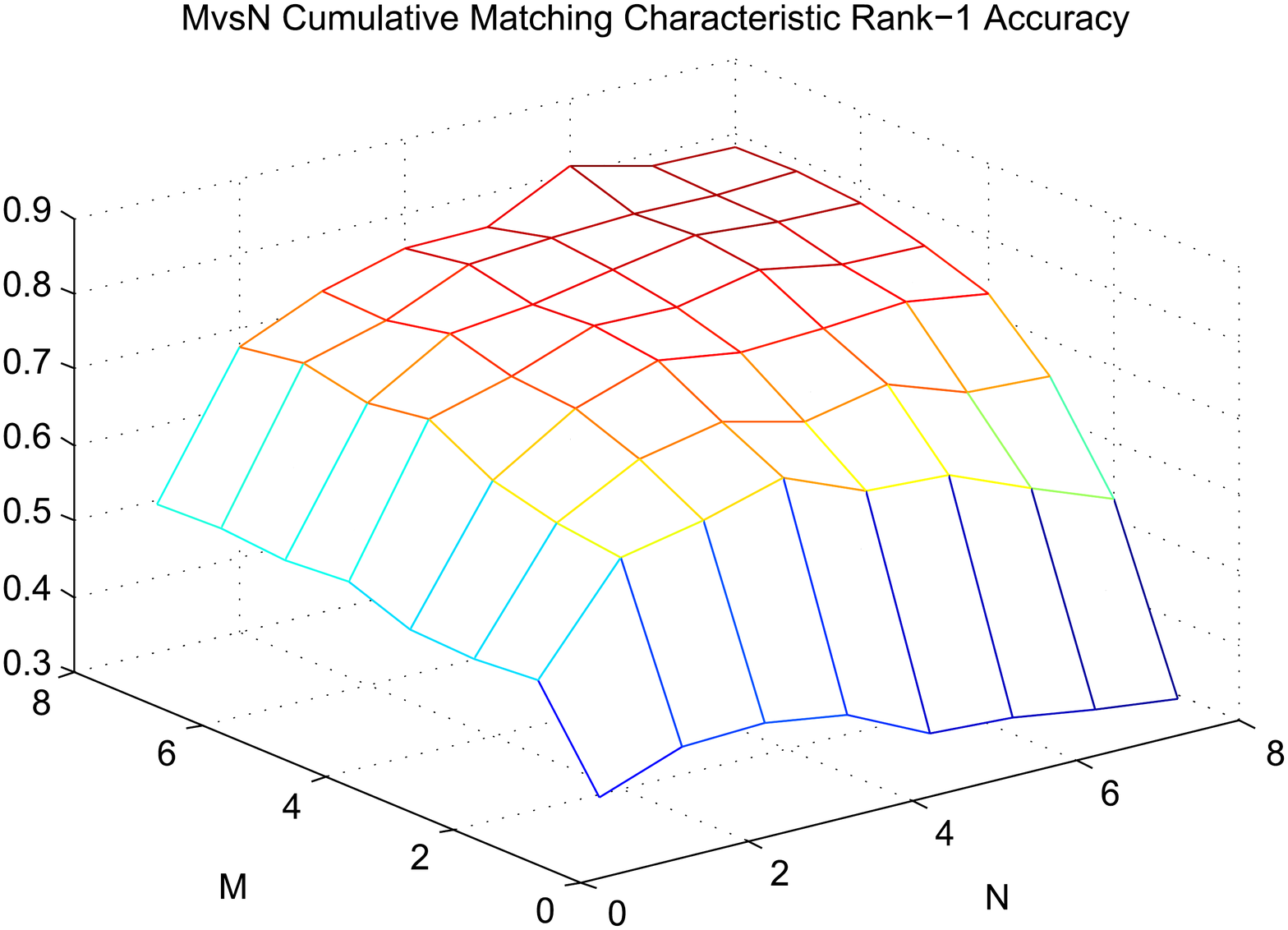}}
  \centerline{(b) Mid-Pooling on PKU-Reid}
\end{minipage}
\hfill
\begin{minipage}[b]{.31\linewidth}
  \centering
  \centerline{\includegraphics[width=6cm]{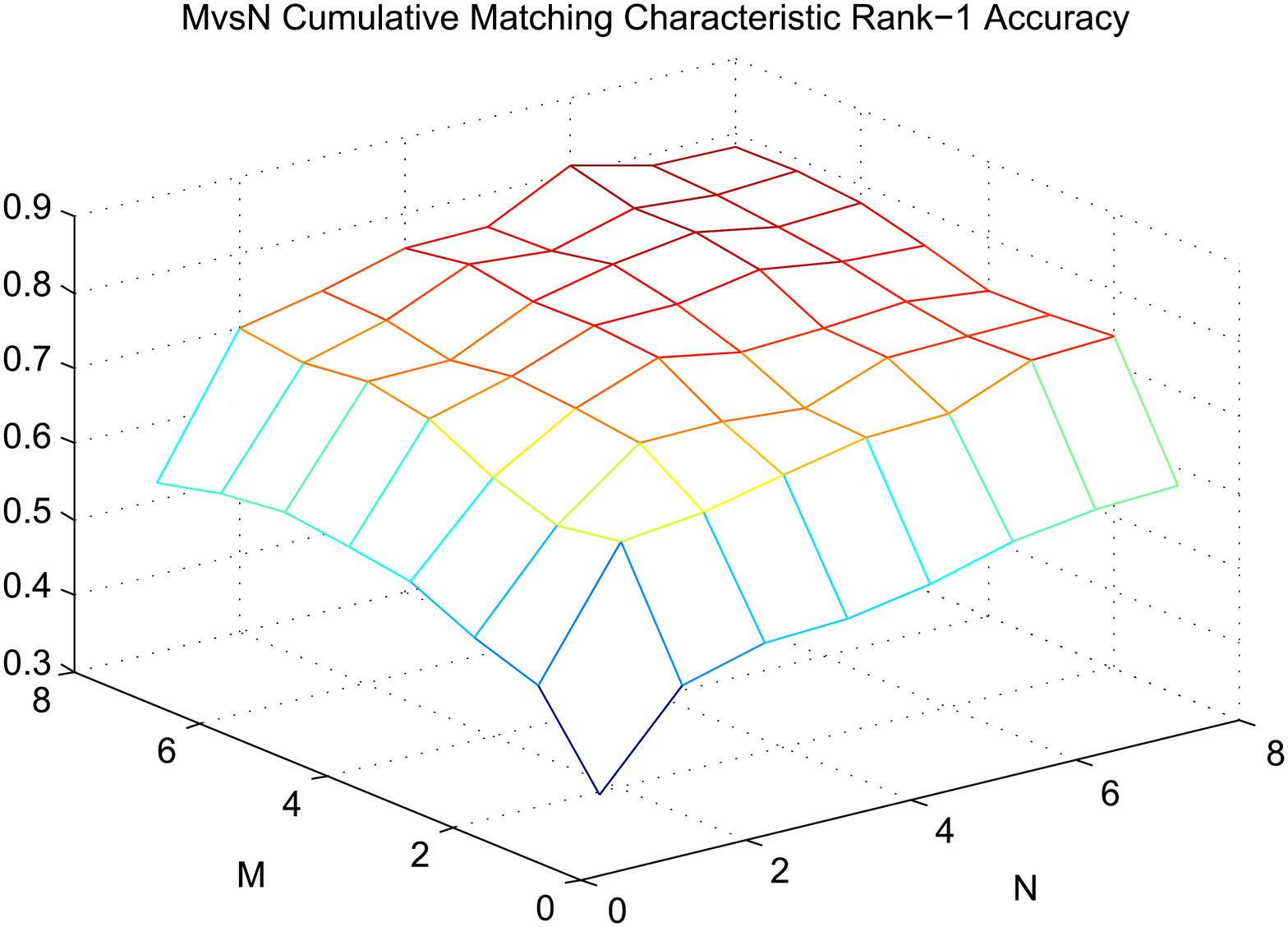}}
  \centerline{(c) ODBoA-Mid-Pooling on PKU-Reid}
\end{minipage}
\caption{Evaluations of ODBoA on PKU-Reid dataset. (a) M probe shots versus M gallery shots CMC rank-1 results for different methods. (b) M probe shots versus N gallery shots CMC rank-1 results for Mid-Pooling method. (c) M probe shots versus N gallery shots CMC rank-1 results for ODBoA-Mid-Pooling method. \textbf{Best viewed in color.}}\label{fig_evaOnPKU-Reid}
  \vspace{-3mm}
\end{figure*}

\vspace{-3mm}
\subsection{Performances of ODBoA}
The helpfulness of person orientation information is evaluated on PKU-Reid dataset.
Performances of person orientation estimation are evaluated on PKU-Reid and TUD Multiview Pedestrian datasets. Performances of ODBoA are evaluated on PKU-Reid, Market-1203 and 3DPeS datasets. 
The dimension of feature vectors is reduced to 80 for PKU-Reid and Market-1203 datasets, and 50 for 3DPeS dataset using gauss kernel PCA empirically. The gauss kernel band width is set to 0.8.

\textbf{Evaluation of person orientation information.}
Towards person re-identification problem, metric learning is widely used for person matching. Traditionally, positive and negative pairs used to train the metric matrix are randomly generated from training data, ignoring the person orientation information. In our experiments, pairs with all orientation, dissimilar orientation, similar orientation and same orientation are generated for positive and negative pairs. Similar orientation includes the same and adjacent orientations, while dissimilar orientation includes the rest. Performances are evaluated on PKU-Reid dataset with ODBoA-Avg (explained in the following experiments).

As depicted in Fig. \ref{fig_oriEva}, using positive pairs of similar orientation for training performs best, and using positive pairs of dissimilar orientation performs worst.
It indicates that orientation information plays an important role in positive pairs generation during metric model training, since appearances obtained in different orientations of one person may vary significantly which introduces some kind of noise. In addition, using positive pairs of same orientation performs a little worse than using similar orientation, especially when few appearances in different orientations are provided. 
It is mainly because of that similar orientation positive pairs provide some appearance variation in training data, which improves the robustness of the metric model.

Besides, orientation information of positive pairs greatly affects the metric model, while that of negative pairs affects little as shown in Fig. \ref{fig_oriEva}. This is because that positive pairs consist of appearances from one person which is sensitive to varying viewpoints. 
However, negative pairs consist of appearances from different persons which is insensitive to varying viewpoints, because the appearances themselves contain large variations.

Therefore, similar orientation positive pairs and all orientation negative pairs are used to train the metric model, $i.e.$, the metric matrix, in the following experiments on PKU-Reid dataset. However, since the appearances with various orientations are insufficient in Market-1203 and 3DPeS dataset, all orientation positive and negative pairs are used in the experiments on these two datasets.

\textbf{Evaluation of person orientation estimation.}
For both PKU-Reid and TUD datasets, half images are used for training, and others are used for testing.
Results are illustrated in Fig. \ref{fig_oriEstimation}, where the Accuracy1 denotes that the result is correct when the predicted and true orientation are same, and Accuracy2 denotes that the result is correct when the predicted and true orientation are same or adjacent.
Since person appearances obtained in adjacent orientations are similar, Accuracy2 evaluation criterion is more suitable for person re-identification problem.
Using only appearance information and baseline method, the Accuracy2 achieves 97.6\% on PKU-Reid and 80.3\% on TUD dataset, respectively.

The possible reason of performance gap in these two datasets is that TUD dataset collects images from a variety of complex scenes while PKU-Reid dataset collects images from two camera views.
When motion information is available, person orientation could be estimated more accurately\cite{AVSS11,Online_Ori}, which contributes to re-identification accuracy.

In order to analyze the influence of person orientation estimation accuracy on person re-identification, we compare the performances of ODBoA-Avg (explained in the following experiments) on PKU-Reid dataset. As shown in Fig. \ref{fig_ori4reid}, more orientation estimation accuracy, $i.e.$, with true orientation, leads to better re-identification performance.
In the rest experiments, we use the true orientation in testing stage for fair comparison.

\textbf{Evaluation of ODBoA.}
PKU-Reid dataset collects person images from all eight orientations to fully evaluate the use of orientation information for person re-identification problem.
However, it is too idealized to capture images in all eight orientations. In this paper, a more realistic and larger dataset, Market-1203 dataset, is constructed to evaluate the effectiveness of ODBoA in practical scenes.
For both datasets, we randomly select half data for training and the rest for testing.
Besides, in testing stage, we use $M$ vs $N$ comparison, which means each person has $M$ shots in probe set and $N$ shots in gallery set. While in training stage, all training data is used.
In the following experiments on PKU-Reid and Market-1203 datasets, we mainly focus on the influence of orientation information.

Here, some experiment settings are declared first:
\begin{itemize}
\item \textbf{Low-Pooling} pools the low-level features of all shots into one signature, and then calculates the similarity between two signatures, namely low-level feature fusion without orientation information.
\item \textbf{ODBoA-Low-Pooling} pools the low-level feature of selected shots into one signature based on orientation, and then calculates the similarity between two signatures, namely low-level feature fusion with orientation information.
\item \textbf{Mid-Pooling} pools the mid-level feature representations of all shots into one signature, and then calculate the similarity between two signatures, namely mid-level feature fusion without orientation information.
\item \textbf{ODBoA-Mid-Pooling} pools the mid-level feature representations of selected shots into one signature based on orientation, and then calculate the similarity between two signatures, namely mid-level feature fusion with orientation information.
\item \textbf{Avg} calculates the average similarity score of each shot pairs, namely decision level fusion without orientation information.
\item \textbf{ODBoA-wAvg} calculates the weighted average similarity score based on orientation, namely decision level fusion with orientation information.\textcolor[rgb]{0.00,0.07,1.00}{ And the features obtained from the images in the same orientation are not fused in feature level.} In our experiments, we use weight 1, 0.9, 0.4 for the images of same, adjacent and other orientation, respectively. These weight parameters are empirical values.
\item \textbf{Dual-Avg} \cite{ICPR14_Ori_Re-id}, the state-of-the-art method, trains the metric models for person in similar and dissimilar orientations, respectively. Average similarity score of each shot pairs is calculated with different metric model depending on orientation information. 
    For fair comparison, we adopt the same feature extraction process and metric model used in our framework for this method.
\item \textbf{Dual-wAvg}, an improved version of Dual-Avg, calculates the weighted average similarity score based on orientation. The weight parameters are also empirical values, $i.e.$, 1, 0.9, 0.4 for the images of same, adjacent and other orientation.
\end{itemize}

\begin{figure}[t]
\vspace{-0.2mm}
\begin{minipage}[b]{1\linewidth}
  \centering
  \centerline{\includegraphics[width=8.5cm]{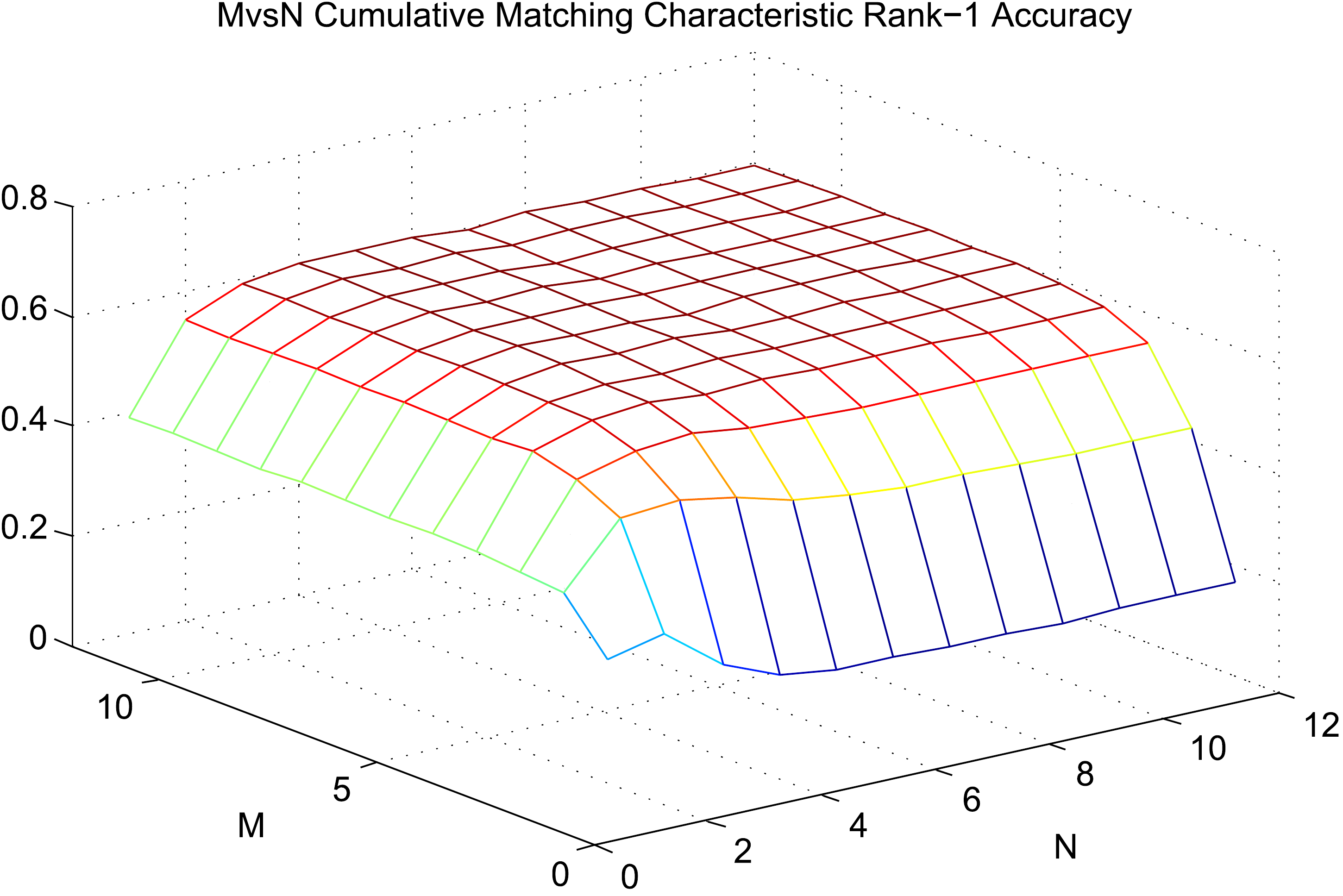}}
  \centerline{(a) Mid-Pooling on Market-1203}\medskip
\end{minipage}
\vspace{-4mm}
\begin{minipage}[b]{1\linewidth}
  \vspace{2mm}
  \centering
  \centerline{\includegraphics[width=8.5cm]{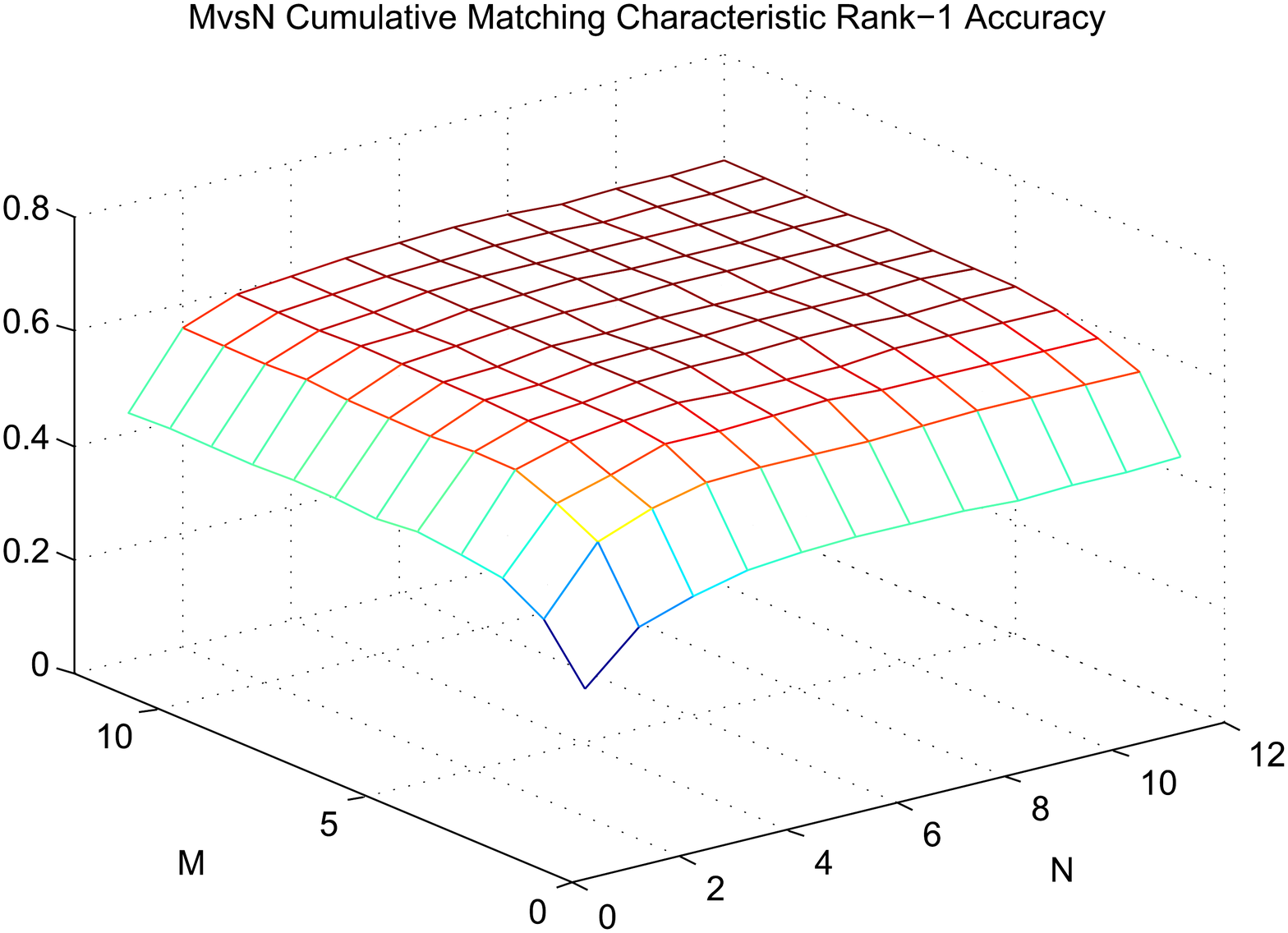}}
  \centerline{(b) ODBoA-Mid-Pooling on Market-1203}\medskip
\end{minipage}
\vspace{-4mm}
\caption{Evaluations of ODBoA on Market-1203 dataset. (a) M probe shots versus N gallery shots CMC rank-1 results for Mid-Pooling method. (b) M probe shots versus N gallery shots CMC rank-1 results for ODBoA-Mid-Pooling method.}\label{fig_evaOnMarket1203}
\end{figure}

As illustrated in Fig. \ref{fig_evaOnPKU-Reid}, ODBoA based fusion methods perform better than non-ODBoA ones, which means that orientation information is very helpful to multi-shot person re-identification.
Comparing the fusion methods in different levels, mid-level feature fusion (red line) performs best. The reasonable explanation is that low-level feature fusion is very sensitive to space misalignment and complex background noise, and decision level fusion could not handle the redundancy and difference between different shots well. However, the proposed mid-level feature fusion approach can deal with these problems well due to the strong representation ability of BSFR and feature selection ability of max pooling. 
Besides, our ODBoA-Mid-Pooling method performs better than the state-of-the-art method Dual-Avg \cite{Trans_Ori_Re-id} and its improved version Dual-wAvg, for our method makes full use of orientation information.
In addition, there is a big improvement from $1$ vs $1$ to $2$ vs $2$, which verifies the data missing problem that appearances obtained in different orientations of one person could vary significantly.

In real surveillance scenes, data imbalance is a common problem that one person may have appearances in only one orientation under camera A and appearances in several orientations under camera B. Data imbalance problem will lead to bad performance for multi-shot person re-identification, especially for $1$ vs $N$ and $M$ vs $1$ as illustrated in Fig. \ref{fig_evaOnPKU-Reid}(b) and Fig. \ref{fig_evaOnMarket1203}(a). 
The information from other orientations may be some kind of noise to the matching, which causes the inaccuracy.
It is verified that orientation information is very helpful to solve the data imbalance problem by comparing Fig. \ref{fig_evaOnPKU-Reid}(c) to Fig. \ref{fig_evaOnPKU-Reid}(b) and comparing Fig. \ref{fig_evaOnMarket1203}(b) to Fig. \ref{fig_evaOnMarket1203}(a).
It is noted that most individuals of Market-1203 have less than six shots under one camera view and many shots are in similar orientations, so the main concentration of improvement with multi-shot is between $1$ vs $1$ and $4$ vs $4$.
Besides, there is an interesting phenomenon that the accuracy matrixes of Mid-Pooling are not symmetric as shown in \ref{fig_evaOnPKU-Reid}(b) and Fig. \ref{fig_evaOnMarket1203}(a). The reasonable explanation is that much noise from complex background and appearances of different orientations is involved into all gallery individuals, when probe contains few shots and gallery contains many shots, $e.g.$, $1$ vs $N$. However, such noise is solely involved into one individual in probe, when probe contains many shots and gallery contains few shots, $e.g.$, $M$ vs $1$.
In conclusion, we fuse multi-shot information with mid-level feature based on orientation information, namely ODBoA-Mid-Pooling.

\begin{table}[t]
\centering
\caption{Comparisons with the state-of-the-arts on 3DPeS}\label{table3}

\begin{tabular*}{8.5cm}
{@{\extracolsep{\fill}} c c c c c c}
\toprule[2pt]
3DPeS & Rank 1 & Rank 5 & Rank 10 & Rank 20 & Rank 50 \\
\midrule[1pt]
RWACN\cite{RWACN} & 41.5 & 65.7 & 74.1 & 83.7 & 95.9 \\
SDALF\cite{SDALF} & 26.2 & 46.1 & 59.5 & 71.6 & 93.6  \\
SoF2\cite{Trans_Ori_Re-id} & 46.9 & 73.3 & 82.4 & 89.8 & 97.0  \\
LMNN-R\cite{LMNN-R} & 23.0 & 44.9 & 55.2 & 69.0 & 88.9  \\
KISSME\cite{KISSME} & 22.9 & 49.0 & 62.2 & 76.0 & 93.2  \\
LF\cite{LFDA} & 33.3 & 58.2 & 70.0 & 81.1 & 95.1  \\
Dual\cite{ICPR14_Ori_Re-id} & 52.6 & - & 82.6 & 91.0 & 96.3  \\
\midrule[1pt]
\textbf{ODBoA}(Ours) & \textbf{55.8} & \textbf{79.0} & \textbf{87.9}& \textbf{93.6} & \textbf{97.5}\\
\bottomrule[2pt]
\end{tabular*}
\end{table}

\textbf{Comparison with state-of-the-arts.}
Comparing experiments of our ODBoA method, $i.e.$, ODBoA-Mid-Pooling (explained in the previous experiments) and the state-of-the-art methods are conducted on 3DPeS dataset.
Table \ref{table3} shows that ODBoA clearly outperforms the other methods.
To illustrate this point, ODBoA has more competitive advantage over the latest method Dual\cite{ICPR14_Ori_Re-id} with best performance on 3DPeS dataset.
The reasonable explanation is that previous methods use little or partial body structure information, while ODBoA makes full use of body structure information including vertical and horizontal directions. Meanwhile, ODBoA uses mid-level feature representation and fusion which are insensitive to space misalignment and robust to the variations of poses and viewpoints.
Note that persons in 3DPeS dataset are captured multiple times not only with different viewpoints, but also at different time instants and on different days, in clear light and in shadow areas. 
Our high performance indicates that our method is more robust to complex scenarios and suitable for practical applications.

\section{Conclusions and feature work}
This paper originally introduces body-structure based feature representation (BSFR) and orientation
driven bag of appearances (ODBoA) for person re-identification. BSFR makes full use of body structure information from horizontal direction by applying the novel body-structure pyramid in both codebook learning and feature pooling steps. ODBoA utilizes the body structure information from vertical direction by integrating person orientation into multi-shot metric model. The proposed framework consisting of BSFR and ODBoA can handel the space misalignment and data missing problem well even for images with complex scenes and inter-class ambiguities.
Experimental results show that our approach can achieve better performance than the state-of-the-art methods and deal with the data imbalance problem well.

In future work, we plan to investigate multi-target tracking algorithms and integrate it with our person re-identification framework to build a integrated intelligent surveillance system. Tracking and re-identification are auxiliary to each other, since tracking could provide new person image sequences online for updating the re-identification model, and re-identification could solve the long time occlusion and appearance change problem in tracking.

\ifCLASSOPTIONcaptionsoff
  \newpage
\fi



%
%
%

\bibliographystyle{./IEEEtran}
\bibliography{./bare_jrnl}

\end{document}